\pdfoutput=1

\documentclass[11pt]{article}

\usepackage[]{latex/acl}

\usepackage{times}
\usepackage{latexsym}

\usepackage[T1]{fontenc}

\usepackage[utf8]{inputenc}

\usepackage{microtype}

\usepackage{inconsolata}

\usepackage{color}

\usepackage{graphicx}
\usepackage{multirow}

\usepackage{multirow}
\usepackage{subcaption}
\usepackage{booktabs}
\usepackage{MnSymbol}
\usepackage{graphicx}
\usepackage{enumitem}
\usepackage{amsmath}
\usepackage{dsfont}
\usepackage{mathtools}
\usepackage{xcolor}
\usepackage{pifont}

\newcommand{\cmark}{\ding{51}}
\newcommand{\xmark}{\ding{55}}

\title{General Purpose Verification for Chain of Thought Prompting}

\author{Robert Vacareanu$^{2,3}$\thanks{~~Work done during internship at AWS AI Labs} ~~~~Anurag Pratik$^1$  ~~~~ Evangelia Spiliopoulou$^1$ ~~~~ Zheng Qi$^1$\\ \textbf{Giovanni Paolini$^4$}\thanks{~~Work done while at AWS AI Labs} ~~~~ \textbf{Neha Anna John$^1$}  ~~~~ \textbf{Jie Ma$^1$}  ~~~~ \textbf{Yassine Benajiba$^{1}$} ~~~~ \textbf{Miguel Ballesteros$^1$}
 \\  $^1$AWS AI Labs \\ $^2$University of Arizona, Tucson, AZ, USA\\ $^3$Technical University of Cluj-Napoca, Romania \\ $^4$Department of Mathematics, University of Bologna, Italy \\
 {\tt \footnotesize{\{anuragik, spilieva, zhengqii, nehajohn, jieman, benajiy, ballemig\}@amazon.com}}
\\ {\tt \footnotesize{g.paolini@unibo.it}} 
\\ {\tt \footnotesize{rvacareanu@arizona.edu}} 
}

\begin{document}
\maketitle
\begin{abstract}
Many of the recent capabilities demonstrated by Large Language Models (LLMs) arise primarily from their ability to exploit contextual information. In this paper, we explore ways to improve reasoning capabilities of LLMs through (1) exploration of different chains of thought and (2) validation of the individual steps of the reasoning process. We propose three general principles that a model should adhere to while reasoning: (i) Relevance, (ii) Mathematical Accuracy, and (iii) Logical Consistency. We apply these constraints to the reasoning steps generated by the LLM to improve the accuracy of the final generation. The constraints are applied in the form of verifiers: the model itself is asked to verify if the generated steps satisfy each constraint. To further steer the generations towards high-quality solutions, we use the perplexity of the reasoning steps as an additional verifier. We evaluate our method on 4 distinct types of reasoning tasks, spanning a total of 9 different datasets. Experiments show that our method is always better than vanilla generation, and, in 6 out of the 9 datasets, it is better than best-of N sampling which samples N reasoning chains and picks the lowest perplexity generation.
\end{abstract}

\section{Introduction}
\begin{figure}[t]
    \includegraphics[width=1.0\columnwidth]{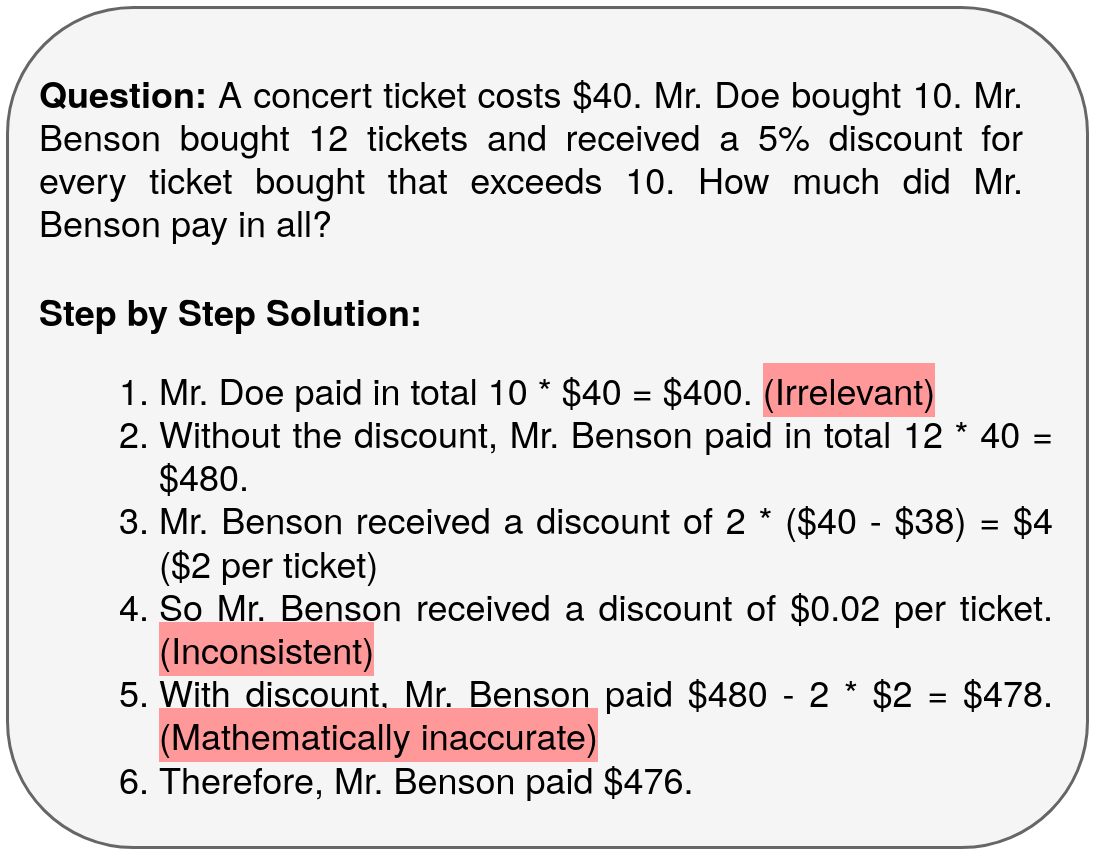}
      \caption{An illustrative example highlighting how the final answer can ultimately be correct (i.e. $12 * \$40 - (12 - 10) * \$40 * 0.05 = \$476$), but it is reached through steps that are (i) irrelevant (Step 1), (ii) contradicting previous steps (Step 4 contradicts Step 3), or (iii) with mathematical errors (Step 5).}
    \label{fig:unfaithful_cot_reasoning_example}
\end{figure}

Large Language Models (LLMs) have demonstrated impressive capabilities of performing a diverse range of tasks by framing them as text generation \cite[\textit{inter alia}]{Brown2020LanguageMA, Chowdhery2022PaLMSL, Touvron2023LLaMAOA, OpenAI2023GPT4TR, Bubeck2023SparksOA}. Chain-of-Thought prompting \cite{Nye2021ShowYW, Wei2022ChainOT, Chowdhery2022PaLMSL} further improved their performance on challenging reasoning tasks using a simple trick of generating intermediate steps before giving the final answer allowing the LLM to spread computation over more tokens \cite{Goyal2023ThinkBY}. 
However, this approach lacks a mechanism to rectify errors in reasoning. While LLMs may eventually reach the correct answer, they might do so via incorrect intermediate reasoning steps, or worse, never reach the correct answer due to earlier mistakes \cite{Turpin2023LanguageMD}.
To illustrate this, we provide a concrete example in Figure~\ref{fig:unfaithful_cot_reasoning_example}, where the final answer is correct, but the intermediate steps are (i) irrelevant \cite{Shi2023LargeLM}, (ii) contradicting previous steps \cite{Mndler2023SelfcontradictoryHO}, and (iii) with mathematical errors \cite{patel-etal-2021-nlp}. 
Recent work \cite{Yao2023TreeOT, Xie2023SelfEvaluationGB, Pan2023AutomaticallyCL} has attempted to alleviate these problems by employing a search mechanism or a self-correction mechanism in the spirit of "System 2" thinking. Other directions include training a dataset-specific verifier to improve the performance when aggregating multiple reasoning chains \cite{li-etal-2023-making}.
However, all these approaches have dataset-specific adaptations and don't generalize out-of-the-box. 

In this work, we explore if catching early mistakes in reasoning chains through problem-agnostic verification can improve reasoning in LLMs. We propose three general principles that a model should adhere to while reasoning: (i) Relevance, (ii) Mathematical Accuracy, and (iii) Logical Consistency and use models, called verifiers, to test for each principle. Each verifier operates on a step \cite{Uesato2022SolvingMW} generated from the step-by-step manner of Chain-of-Thought prompting and assigns a score to that step. We design the verifiers to operate at this granularity so they can detect intermediate mistakes and discourage the LLM from committing to an erroneous reasoning chain. To further steer the generation towards better steps, we use the perplexity of the reasoning step as an additional verifier. We then explore various ways, including Self-Consistency \cite{Wang2022SelfConsistencyIC}, to aggregate verifier scores and report their downstream task performance.

We make the following contributions: (i) we propose a general framework for guiding reasoning in LLMs using verifiers which offers the flexibility to use a problem-agnostic implementation across any reasoning task but also offers the adaptability to use task- and dataset-specific implementations, and
(ii) we show how using our proposed verifiers can improve reasoning outcomes in LLMs and can also improve existing ensembling techniques like Self-Consistency.  
Importantly, our work is not intended to be an exploration on the best way to use a computational budget to achieve a desired performance, but an exploration of whether the LLM are capable (even if inefficiently) of detecting their own mistakes together with a simple recovering mechanism.

\section{Related Work}

We focus here only on LLM-based approaches, and divide previous related work according to
(i) the generalizability of the prompts used, and 
(ii) how the final answer is generated.

\paragraph{Types of Prompts}
In prior work, the prompts used can be categorized based on their level of generality. Some approaches utilize a singular prompt, applying it uniformly across a wide spectrum of datasets and tasks.
\citet{Wei2022ChainOT} proposed chain-of-thought prompting with in-context examples. 
\citet{Kojima2022LargeLM} then explored zero-shot prompts capable of exhibiting similar behaviors. 
Other recent works explore using LLMs to self-evaluate \cite{Yin2023DoLL} and potentially improve upon their generation with the resulting feedback \cite{Saunders2022SelfcritiquingMF, Chen2023TeachingLL, Pan2023AutomaticallyCL, Shinn2023ReflexionLA}. 
\citet{Bai2022ConstitutionalAH} use an LLM with in-context examples to detect and edit the responses of a chat model that are harmful or toxic. 
\citet{Madaan2023SelfRefineIR} proposes a framework to iteratively self-improve the generations of a LLM. 
\citet{Yao2023TreeOT} tightly integrates an LLM with custom dataset-specific prompts to act as a guiding mechanism in the underlying search space. \citet{Hao2023ReasoningWL} expands on this by using a Monte Carlo tree search strategy.
Other recent work questioned the extent to which using an LLM to evaluate and improve its own generations is viable \cite{Huang2023LargeLM}, a conclusion which we observed as well in our preliminary work and sidestepped by re-sampling instead of asking the LLM to refine.

Importantly, previous work explored self-evaluation through the lens of task-specific evaluation and prompts, a direction that inherently constrains the broader utility of Large Language Models (LLMs) as general-purpose reasoners. On the other hand, our approach follows a distinct trajectory: we deliberately eschew the use of prompts tailored to individual datasets or tasks.

\paragraph{How the Final Answer is Generated}
A second dimension is that of how the model arrives at the final solution, where we distinguish between methods that take a linear approach \cite{Wei2022ChainOT, Kojima2022LargeLM, Goyal2023ThinkBY} from the methods that do not \cite{Yao2022ReActSR, Wang2023DescribeEP, Long2023LargeLM}.
By linear approaches, we refer to those methods where the final answer is generated token-by-token in one go. On the other hand, non-linear approaches typically include a search mechanism \cite{Xie2023SelfEvaluationGB, Yao2023TreeOT, Besta2023GraphOT} or a self-reflection process \cite{Madaan2023SelfRefineIR, Pan2023AutomaticallyCL}.

For example, recent work explored tightly integrating the LLM to act as a guiding mechanism in the underlying search space \cite{Xie2023SelfEvaluationGB, Yao2023TreeOT}. This involves one LLM generating candidate steps while another LLM assigns single float value as a value score. This value score is derived from an LLM with a dataset-specific prompt and in-context examples.

Within this dimension, our approach aligns with the non-linear paradigm. We leverage verifiers to evaluate each step in the solution-generation process, with the overarching aim of guiding the generation towards solutions that receive high scores, as determined by the verifiers. Differently from previous work on self-evaluation, we explore a setting of self-evaluation that is problem-agnostic.
\section{Proposed Method}

\begin{figure*}[t]
    \centering
    \includegraphics[width=0.95\textwidth]{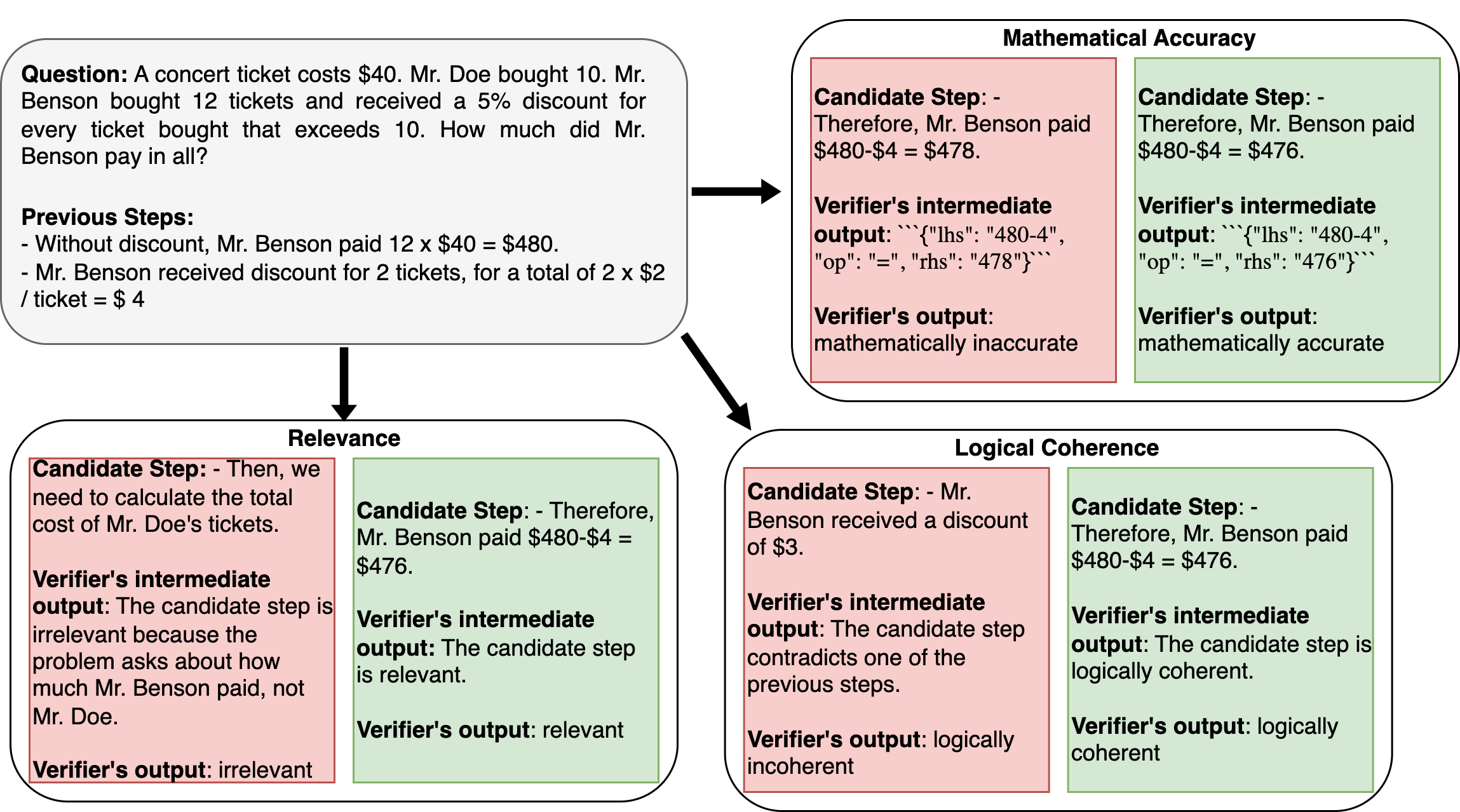}
    \caption{An example of each of our proposed verifiers applied to a given question and previous steps.}
    \label{fig:pm_rmalc_example}
\end{figure*}

We propose a novel approach that seamlessly integrates with any given Large Language Model's (LLM) solution generation process, at both step generation and step evaluation \cite{Wei2022ChainOT, Wang2022SelfConsistencyIC, Kojima2022LargeLM}. Our approach consists of two components: a solution generator $G$ and a set of verifiers $\mathcal{V}$, where each verifier specializes in a particular qualitative aspect of reasoning. The solution generator $G$ is responsible for generating candidate steps, and each verifier $v \in \mathcal{V}$ is responsible for checking whether the candidate step is in compliance with the specific reasoning property. We explore properties that are generally applicable to a wide range of reasoning tasks. We provide an illustrative example in Figure~\ref{fig:pm_rmalc_example}.

This section is further organized as follows. We describe the notations and abstractions used in Section~\ref{ss:notation}, the solution generation component in Section~\ref{ss:solution_generation}, the proposed verifiers in Section~\ref{ss:step_verification}, the procedure to obtain fine-grained scores for reasoning chains in Section~\ref{ss:css} and how we use verifiers in aggregate in Section~\ref{ss:aggregate_rc}.

\subsection{Notation}
\label{ss:notation}

In the following, we define the notation we adopt throughout the paper.

A token as $t \in \texttt{Vocab}$ where \texttt{Vocab} represents the set of possible tokens defined by a given vocabulary. We use $[t_1, \dots, t_n] \in T$ to represent a given text, with $T$ denoting the set of all potential texts of varying lengths. Under this notation, we represent a problem as $q \in T$ and a reasoning step as $r \in T$, both presented in free-text form. A solution generator $G$ in the form of a function that takes text as input and returns text as output: $G : T \rightarrow T$. We interpret the output text as a sequence of reasoning steps $R^q = [r_1^q, \dots, r_n^q]$. For simplicity, we define a reasoning step as the sequence of tokens until a new line, similar to \citet{Uesato2022SolvingMW}. A verifier $v \in \mathcal{V}$ implemented a function that takes text as input and returns an indicator: $v : T \rightarrow \{0,~1\}$. The returned value represents whether the reasoning step satisfies the verifier constraint ($1$) or not ($0$).

We will next describe the Solution Generator and Step Verification components.

\subsection{Solution Generation}
\label{ss:solution_generation}
The solution generator (typically an LLM) operates over a prompt $q$ and generates a sequence of tokens as output: $G : T \rightarrow T$. 
For our purpose, we concentrate on the correctness at the level of a reasoning step, instead of individual tokens. 
A reasoning step, as defined in this work, represents the sequence of tokens up to the occurrence of a new line \cite{Uesato2022SolvingMW}.
The next reasoning step can then be generated by conditioning the generator $G$ on both the question $q$ and on the sequence of previously generated reasoning steps $R_{1:i}$, which is initially empty. This conditioning can be expressed as follows: $P(r^q_{i+1} | q, R^q_{1:i})$.

For the solution generation process, we adopt the zero-shot prompt used in \citet{Kojima2022LargeLM} which simply appends "Let's think step by step" to the problem question to elicit a chain-of-thought like behavior in the model's response without any annotated exemplars \cite{Wei2022ChainOT}. Nevertheless, our proposed method is agnostic to the specific implementation of the solution generator.

\subsection{Step Verification}
\label{ss:step_verification}
Our research aims to investigate whether specifying a subset of conditions that an ideal reasoning chain should satisfy and then employing these conditions to score the corresponding reasoning chain can result in improved performance on downstream reasoning tasks.

To this end, we explore three general and necessary (but not sufficient) conditions that a given step should satisfy in order for the resulting solution to be sound from a reasoning perspective: (i) Relevance, (ii) Mathematical Accuracy (if applicable), and (iii) Logical Coherence. In this work, we use for our verifiers a set of LLMs provided with a detailed instruction of the task and constraints.\footnote{Prompts available in Appendix~\ref{appendix:verifiers}}
We then map the output of the LLM to $\{0, 1\}$ based on its content. If the relevance verifier generates ``not relevant'', for example, we interpret this as a score of $0$.
Importantly, to verify the generalizability of our proposed methodology, we keep the implementation of our verifiers fixed for all reasoning tasks and for all datasets. We provide an illustrative example of the verifiers we use in Figure~\ref{fig:pm_rmalc_example}. 
We provide in-context exemplars for the Mathematical Accuracy verifier due to challenges in the LLM's ability to generate a valid intermediate structured output, a requirement of this verifier's setup.\footnote{We use the same in-context exemplars for all the Mathematical datasets.} 
In addition to the scores from the aforementioned three verifiers, we use the perplexity score of a reasoning step as an additional verifier, in order to encourage the final solution towards text deemed more likely by the LLM. 
We explain each verifier in greater detail below.

\subsubsection{Relevance}
\label{sss:v_relevance}
The first verifier in our proposed framework is the Relevance verifier, where the goal is to constrain a reasoning step to contribute to the construction of a meaningful solution narrative. We provide a \textit{relevant} and an \textit{irrelevant} example in Figure~\ref{fig:pm_rmalc_example}. 
We begin with a question, a sequence of previous steps, and a candidate step. The Relevance verifier assesses the candidate reasoning steps for their relevance to the problem at hand. In the example provided, calculating how much Mr. Doe spent is irrelevant because the problem asks for how much Mr. Benson spent and there is no connection between them.

We acknowledge the inherent subjectivity and nuance associated with determining the relevance of a given reasoning step. However, there are instances where it becomes evident that a reasoning step is distinctly irrelevant, deviating from the coherent solution narrative. For instance, some reasoning steps may veer into speculative or unrelated content, which our Relevance verifier aims to identify.

\subsubsection{Mathematical Accuracy}
\label{sss:v_mathematical_accuracy}
The Mathematical Accuracy constraint enforces the need for each reasoning step to contain correct mathematical calculations. We implement this in a similar manner to Tool-based approaches \cite{Schick2023ToolformerLM}, working as follows. First, we extract the mathematical formulas (if present) from a sentence as structured output, as depicted in the \textit{Verifier's intermediate output} field corresponding to the Mathematical Accuracy constraint in Figure~\ref{fig:pm_rmalc_example}. For each mathematical calculation present, we extract the left-hand side (\textit{lhs}), the right-hand side (\textit{rhs}), and the operator (\textit{op}).
Then, we programmatically execute the extracted formulas (if any) and compare them using the extracted operator.

\subsubsection{Logical Consistency}
\label{sss:v_logical_consistency}
A third condition for a logically sound argument we use is for the reasoning steps to not contradict each other \cite{M1941TheBW}. To this end, we introduce Logical Consistency as our third verifier.
This verifier operates over the previous steps and the current candidate step.
For example, in Figure~\ref{fig:pm_rmalc_example}, the candidate step \textit{- Mr. Benson received a discount of \$3.} contradicts one of the previous steps, as one of the previous steps already established that Mr. Benson received a discount of \$4.

\subsubsection{Step-wise Perplexity}
\label{ss:v_perplexity}

In addition to the scores resulting from our previously introduced constraint verifiers, we leverage step-wise perplexity as another source of signal, with the goal of favoring lower-perplexity solutions. For each reasoning step $r_i = [t_1, \dots, t_n]$, we compute the perplexity over its token constituents. We hypothesize that lower-perplexity reasoning steps are more desirable, as a lower perplexity prompt is correlated with a higher final performance \cite{Gonen2022DemystifyingPI}. We can interpret a partial reasoning chain $R_{1:i} = [r_1, \dots, r_i]$ as (part of) a prompt that will be used to generate $r_{i+1}$, therefore making the findings in \citet{Gonen2022DemystifyingPI} applicable for our purposes.

\subsection{Constraint Satisfaction Score}
\label{ss:css}

Except for the Perplexity verifier, all our proposed verifiers output a binary value, representing whether a given reasoning step satisfies the given constraint or not. For example, if the Relevance verifier gives a score of $1$ for a given reasoning step $r$, this means that the given step is deemed as relevant.
Since the underlying implementation In order to reduce the variance and get a more fine-grained score $s$, we use the expected value: $s = \mathds{E}(\mathds{1}_{v}(r))
$, which we approximate using sampling. Since each verifier is implemented with an LLM, we can sample multiple generations, map each one to a binary value $\{0, 1\}$, and then average.

\subsection{Using Verifiers in Aggregate}
\label{ss:aggregate_rc}
\subsubsection{Scoring a Reasoning Chain $R$} Given a verifier $v \in V$, we extend the concept of a score for a given reasoning step $r$ to the score for a given (partial or not) reasoning chain $R$ by aggregating the scores over each of its constituent reasoning steps.
Formally, we extend the verifier's scores to that of a reasoning chain $R = [r_1, \dots, r_i]$, where we first obtain a score for each $r_i$, resulting in the following score vector:  
$[\mathds{E}(\mathds{1}_{v(r_1)}), \dots, \mathds{E}(\mathds{1}_{v(r_i)})]$, and then aggregate.
A low-scoring reasoning step does not necessarily render the entire reasoning chain wrong, but it does increase the likelihood of inaccuracies. To combine these scores, we employ the geometric mean as a milder alternative to the \textit{min} operator in our aggregation process.

$$
v(R) = GM([\mathds{E}(\mathds{1}_{v}(r_1)), \dots, \mathds{E}(\mathds{1}_{v}(r_i))])
$$
We obtain a single score for a given reasoning chain $R$ and a set of verifiers $\mathcal{V}$ by aggregating over the scores of each verifier $v \in \mathcal{V}$ on $R$. 
Our proposed framework allows for the customization of each verifier's contribution during aggregation. We use a weighted arithmetic mean, as defined below.

$$
\mathcal{V}(R) = \frac{\sum_{i=1}^{|\mathcal{V}|} w_i \times v_i(R)}{\sum_{i=1}^{|\mathcal{V}|} w_i}
$$

We set $w=2$ for perplexity and $w=1$ for all the others. 
We selected $w=2$ for perplexity based on preliminary experiments on the train partition of GSM8k and CSQA 2.0.
Importantly, we use the same weights for all our experiments.

\subsubsection{Ensembling Methods using the Verifiers}
Ensembling techniques work by aggregating the solution of multiple reasoning chains to obtain a final solution. For example, Self-Consistency \cite{Wang2022SelfConsistencyIC} randomly samples a given number of reasoning chains, and then performs a majority vote on the final answer. 
Instead of resorting to a majority voting mechanism over randomly sampled reasoning chains, we propose to leverage the scores obtained from our proposed verifiers to do the selection and the weighting.

\section{Experiments}

\begin{table*}[t]
    \centering
    \resizebox{1.0\textwidth}{!}{


\begin{tabular}{rrrrrrrrrr}
\toprule
 & \multicolumn{1}{c}{Other} & \multicolumn{3}{c}{Commonsense} & \multicolumn{2}{c}{Symbolic} & \multicolumn{3}{c}{Math} \\
\cmidrule(lr){2-2}\cmidrule(lr){3-5}\cmidrule(lr){6-7}\cmidrule(lr){8-10}
 & BigBench Date & CSQA & CSQA 2.0 & Strategy & Coinflip & Last Letter (2) & GSM8k & SVAMP & AddSub \\
\midrule
Random Chain & 55.77$\pm$3.06 & 47.91$\pm$1.22 & 58.75$\pm$1.68 & 56.03$\pm$0.99 & 58.67$\pm$2.04 & 15.68$\pm$1.51 & 29.23$\pm$1.58 & 38.78$\pm$1.26 & 41.04$\pm$1.79 \\
Low PPL Chain & \underline{63.69$\pm$0.00} & \underline{48.81$\pm$0.00} & \underline{59.10$\pm$0.00} & \textbf{60.90$\pm$0.00} & \textbf{72.80$\pm$0.00} & \textbf{45.00$\pm$0.00} & \underline{40.50$\pm$0.00} & \underline{53.50$\pm$0.00} & \underline{58.48$\pm$0.00} \\
Top Chain wrt Verifiers & \textbf{69.12$\pm$0.21} & \textbf{56.79$\pm$0.12} & \textbf{62.16$\pm$0.22} & \underline{57.21$\pm$0.17} & \underline{64.02$\pm$0.21} & \underline{41.64$\pm$0.44} & \textbf{45.94$\pm$0.30} & \textbf{56.36$\pm$0.23} & \textbf{62.34$\pm$0.25} \\
\bottomrule
\end{tabular}

    }
    \caption{Comparison between two baselines: (1) Random Chains, and (2) Low PPL Chain, and our proposed method, Top Chain wrt Verifiers. In this setting, we record the performance when selecting \textit{one} reasoning chain, according to each method, according to each method's selection criteria. We report Accuracy ($\uparrow$).}
    \label{tab:sct}
\end{table*}

\subsection{Experimental Setting}
\paragraph{Models} We use \textit{Falcon}\footnote{Specifically, we use \textit{Falcon-40B-Instruct}} \cite{Almazrouei2023TheFS} as our base LLM, as it was one of the largest and most capable open-source model family freely available at the time of the experiments.\footnote{Open source according to https://opensource.org/} We use the same model for both solution generation and solution verification. For solution generation, we use the zero-shot prompt from \cite{Kojima2022LargeLM}. For verification, we use different prompts for each verifier. We include the prompts we used in the Appendix~\ref{appendix:verifiers}.

\paragraph{Datasets} We perform experiments spanning 4 reasoning tasks: Math, Commonsense, Symbolic, and Other, and 9 datasets: BigBench Date Understanding \cite{srivastava2023beyond} (\textit{Other)}, CommonsenseQA \cite{talmor-etal-2019-commonsenseqa}, CommonsenseQA 2.0 \cite{Talmor2021CommonsenseQA2E} and Strategy \cite{Geva2021DidAU} (\textit{Commonsense}), Coinflip and Last Letter Concatenation \cite{Wei2022ChainOT} (\textit{Symbolic}), GSM8k \cite{Cobbe2021TrainingVT}, SVAMP \cite{patel-etal-2021-nlp}, and AddSub \cite{Kojima2022LargeLM} (\textit{Math}). We provide an example from each dataset in Appendix~\ref{appendix:datasets}.\footnote{For Last Letter Concatenation, we use only $2$ words instead of $4$.}

We use the standard evaluation metrics as previous work, which is Accuracy score computed between the gold answer and the predicted answer. 

\paragraph{Proposed Method Setting}
All our proposed verifiers are dataset-agnostic and we use the same prompts for all our experiments. Due to computational constraints, we use the mathematical accuracy verifier only for the math datasets. 
In an attempt to minimize the impact of the underlying search strategy for the step-by-step solution, we adopt the following approach: we first sample 40 reasoning chains for each problem and for each dataset, then use the scores resulting from our proposed verifiers to guide our selection process. 

\paragraph{Baselines} We analyze the contributions of the verifiers by comparing the performance of the proposed method against the following baselines: (i) \textit{Random Chains}, where we use the LLM to sample a solution. We use the same prompt as \citet{Kojima2022LargeLM}, and (ii) \textit{best-of N} sampling \cite{adiwardana2020humanlike, Wang2022SelfConsistencyIC}, where we sample a total of $40$ reasoning chains and select the one with the lowest perplexity. Our motivation for using these two baselines is two-fold. First, we want to allow both the baselines and our proposed method to have access to the same candidate reasoning chains. Secondly, it has been observed in \citet{Wang2022SelfConsistencyIC} that Best-of N sampling performs better than greedy decoding, especially for large N.

\paragraph{Experiments}
We conduct the following experiments: (i) Single chain analysis, where we use the resulting scores of each reasoning chain to select a single reasoning chain, (ii) Self-Consistency, where we aggregate multiple reasoning chains, and (iii) Single chain analysis with incomplete chain scoring, where we only score the reasoning chains based on the an initial \% (or number) of the reasoning steps.

\subsection{Single Chain}
\label{ss:single_chain}
In this experiment, we assess how the scores generated by our proposed verifiers are correlated with the likelihood of a reasoning chain reaching the correct final answer.
For this purpose, we conduct the following experiment: from the $40$ sampled reasoning chains, we select the highest-scoring chain based on our proposed verifiers' scores.
We present our results in Table~\ref{tab:sct}, comparing our proposed approach with two baselines: \textit{Random Chains} and \textit{best-of N} sampling.
We make the following remarks.

First, over all datasets, our proposed method performs better than selecting a reasoning chain at random, with improvements ranging from 1.18 points (Strategy) to 25.68 points (Last Letter), with an average improvement of 12.63.

Second, we remark that our proposed method outperforms the best reasoning chain according to perplexity (Low PPL Chains) in over 65\% of the cases ($6$ out of $9$ datasets). This means that our proposed verification procedure provides valuable information beyond what is captured by simply selecting the lowest perplexity chains. 
We note that this trend does not hold true for Symbolic Reasoning, where for both datasets investigated (\textit{Coinflip} and \textit{Last Letter}) the Low PPL Chain is better than the one selected according to our proposed verifier. An exploration over \textit{Coinflip} revealed that steps where the coin has not been flipped received, on average, a lower relevance score, although this information is relevant. We leave the exploration of better verifiers for Symbolic Reasoning to future work.
All in all, the average improvement of our proposed method over the reasoning chain with the lowest perplexity is 1.43 points.

\begin{figure*}[!t]

    \begin{subfigure}[b]{0.31\textwidth}
        \includegraphics[width=1.1\columnwidth]{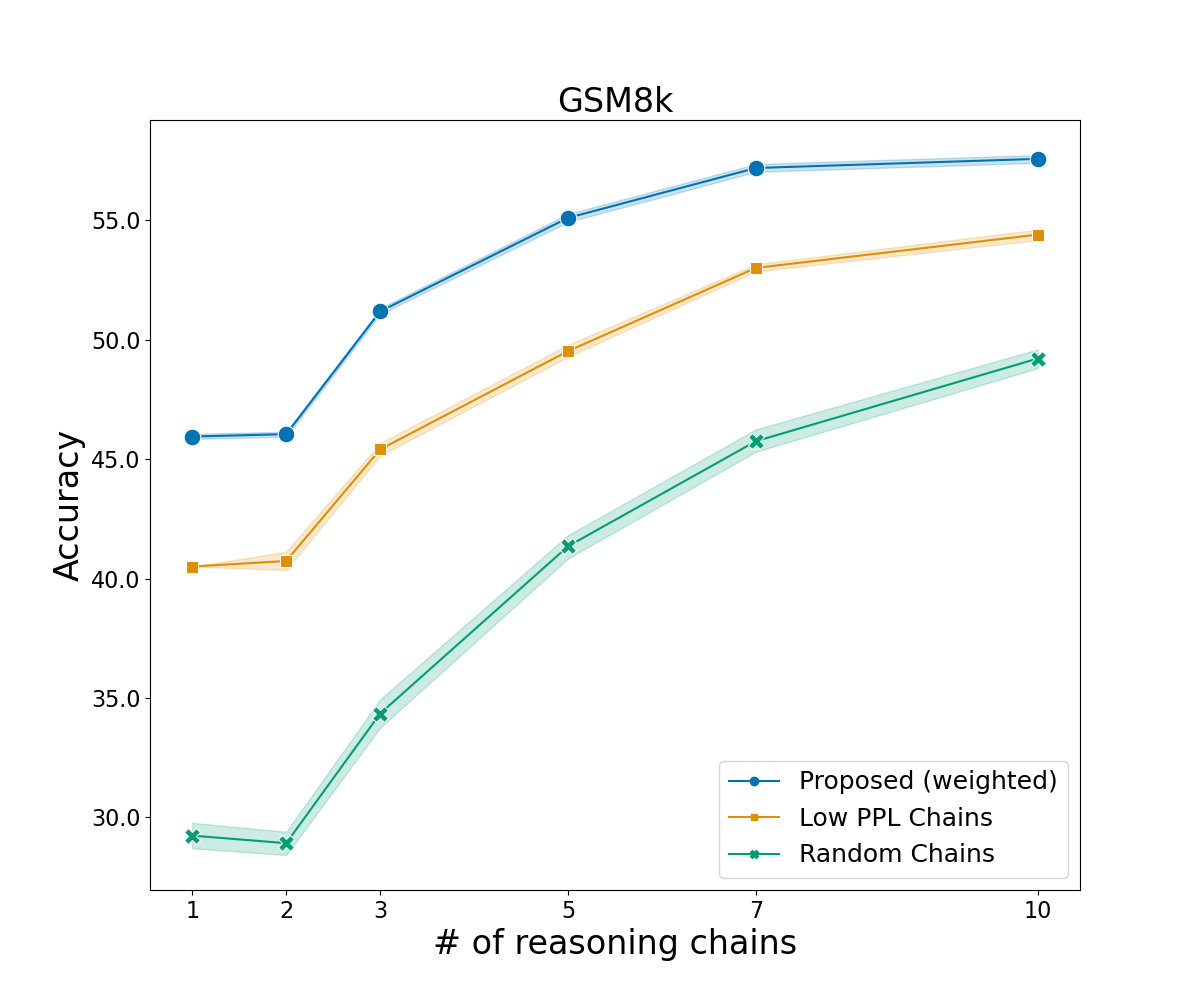}
        \caption{GSM8k}
        \label{l1}
    \end{subfigure}
    \hfill
    \begin{subfigure}[b]{0.31\textwidth}
        \includegraphics[width=1.1\columnwidth]{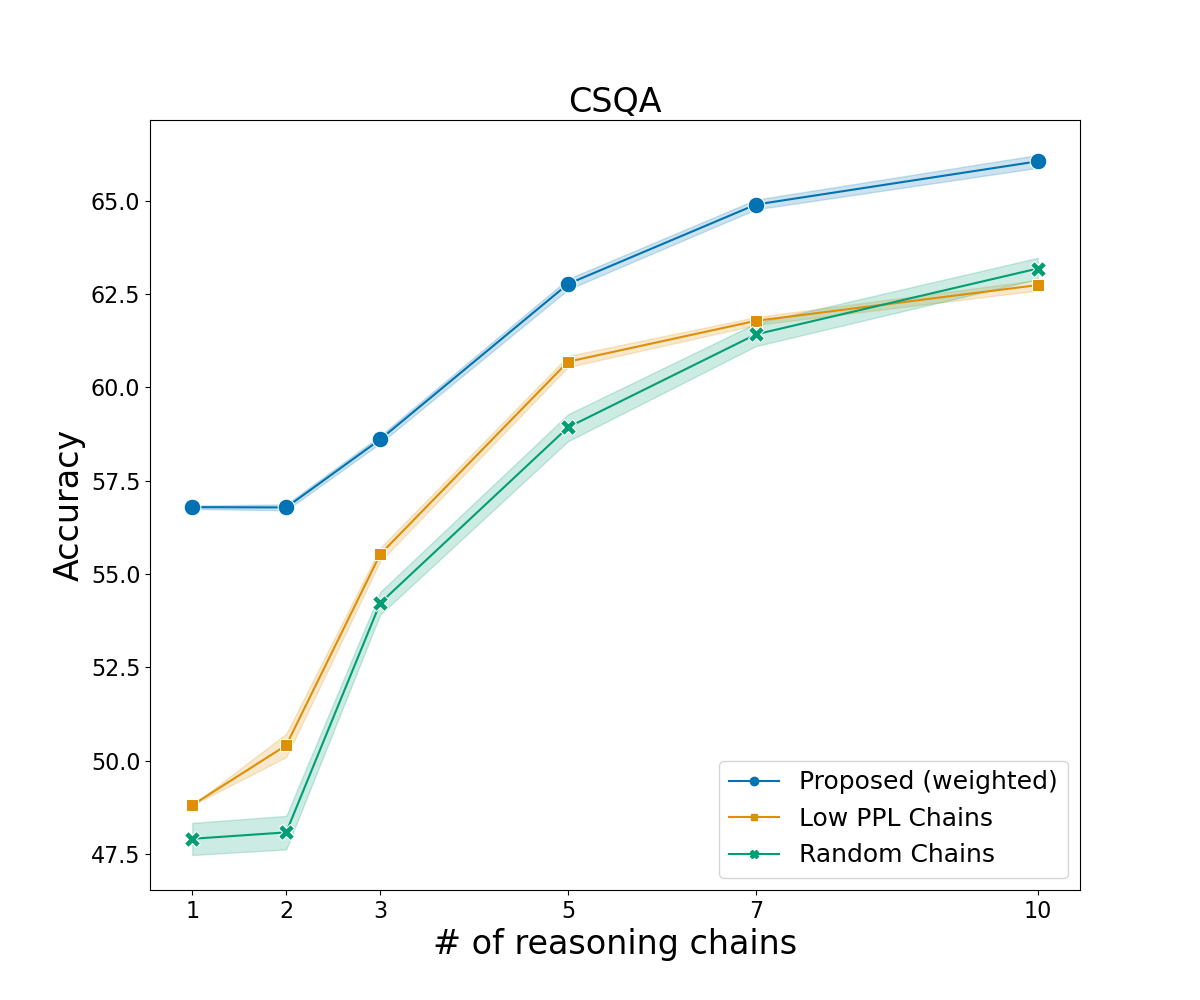}
        \caption{CSQA}
        \label{l1}
    \end{subfigure}
    \hfill
    \begin{subfigure}[b]{0.31\textwidth}
        \includegraphics[width=1.1\columnwidth]{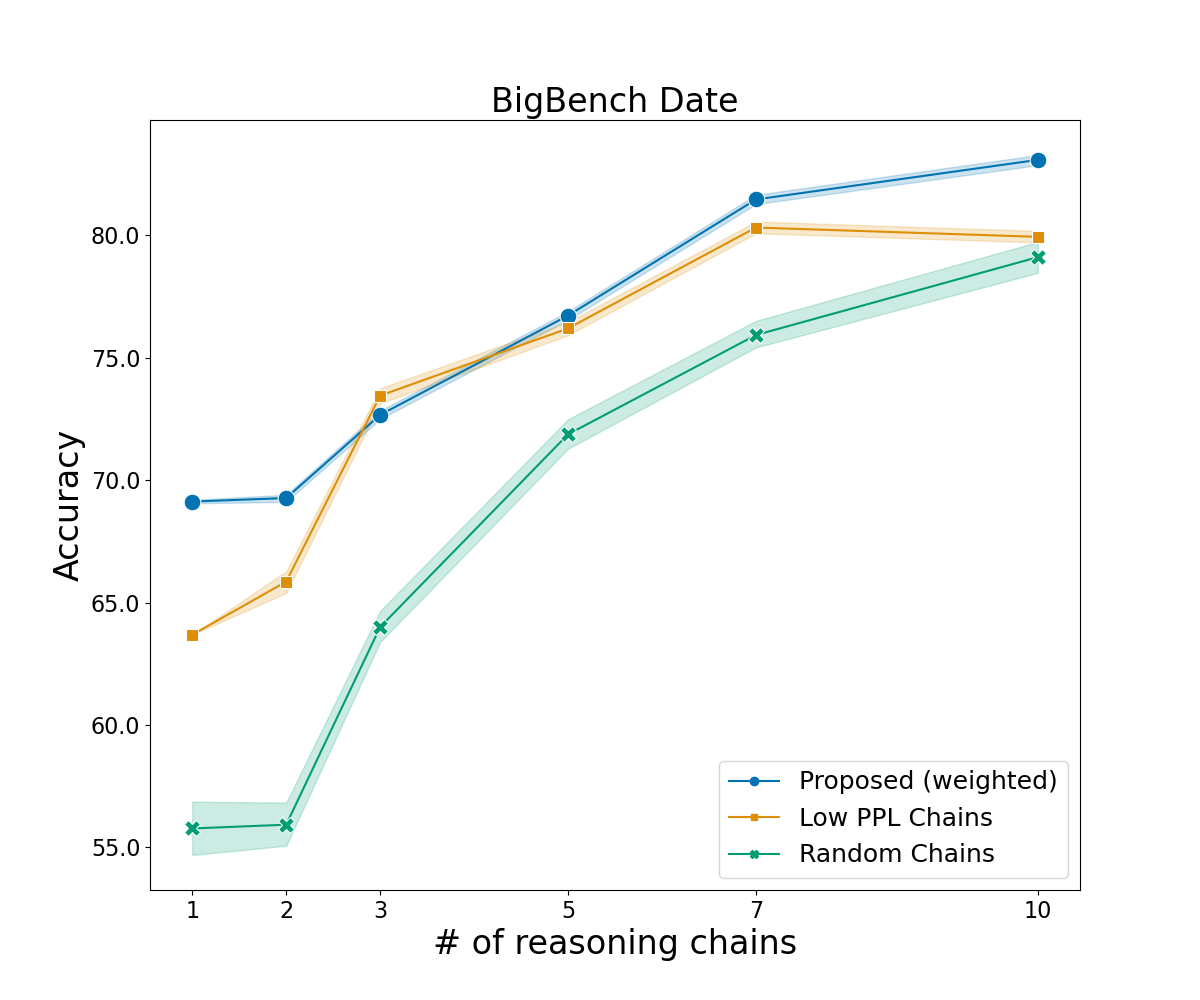}
        \caption{BigBench Date Understanding}
        \label{l1}
    \end{subfigure}
    \caption{Comparison between our proposed method and two baselines, when using Self-Consistency and between 1-10 reasoning chains. We report Accuracy ($\uparrow$).}
    \label{fig:sc_train}
\end{figure*}

\subsection{Self-Consistency}
\label{ss:self_consistency}
In this experiment, we explore how well our proposed method leverages ensemble techniques, particularly Self-Consistency \cite{Wang2022SelfConsistencyIC}. 
We start with the same set of $40$ reasoning paths and employ different selection strategies to evaluate their effectiveness: (i) we randomly sample from these paths (\textit{Random Chains}), (ii) we select chains with the lowest perplexity from this set (\textit{Low PPL Chains}), and (iii) we choose the reasoning chains with high scores, as determined by the verifiers, from these 40 paths (\textit{Proposed (weighted)}).
Unlike the original majority vote approach, we empirically found that weighting each reasoning chain by their verifier scores yields slightly better results. However, it is worth noting that this improvement does not hold when using only perplexity, as shown in \cite{Wang2022SelfConsistencyIC}.
We show in Figure~\ref{fig:sc_train} the behavior of our proposed method. We make two observations: First, our proposed method is able to leverage ensembling techniques, showing consistent performance gains as the number of reasoning chains increases. Second, we remark that our proposed method scales better, consistently outperforming the baselines.

We further investigate the performance impact of weighted voting, utilizing scores from our proposed verifiers, against the standard majority-voting approach. Specifically, we apply both voting methods to the \textit{identical set} of reasoning chains, initially selected at random. 
We found that using the scores of our proposed verifiers to do a weighted voting improves over the majority voting in over 96\% of the cases.\footnote{$78/81$} 
Due to space constraints, we include the resulting plots in Appendix~\ref{appendix:sc_voting}.

\subsection{Verifying Incomplete Reasoning Chains}
\label{ss:incomplete_rc}
In our prior experiments, our proposed verifiers evaluated complete reasoning chains. Now, we explore their effectiveness when applied exclusively to the initial reasoning steps. This experiment provides insights into the potential utility of our proposed method in an "online" setting, where the reasoning step-level evaluation is employed to guide the search for good reasoning chains without fully generating multiple candidate solutions.

\begin{figure*}[!t]

    \begin{subfigure}[b]{0.31\textwidth}
        \includegraphics[width=1.1\columnwidth]{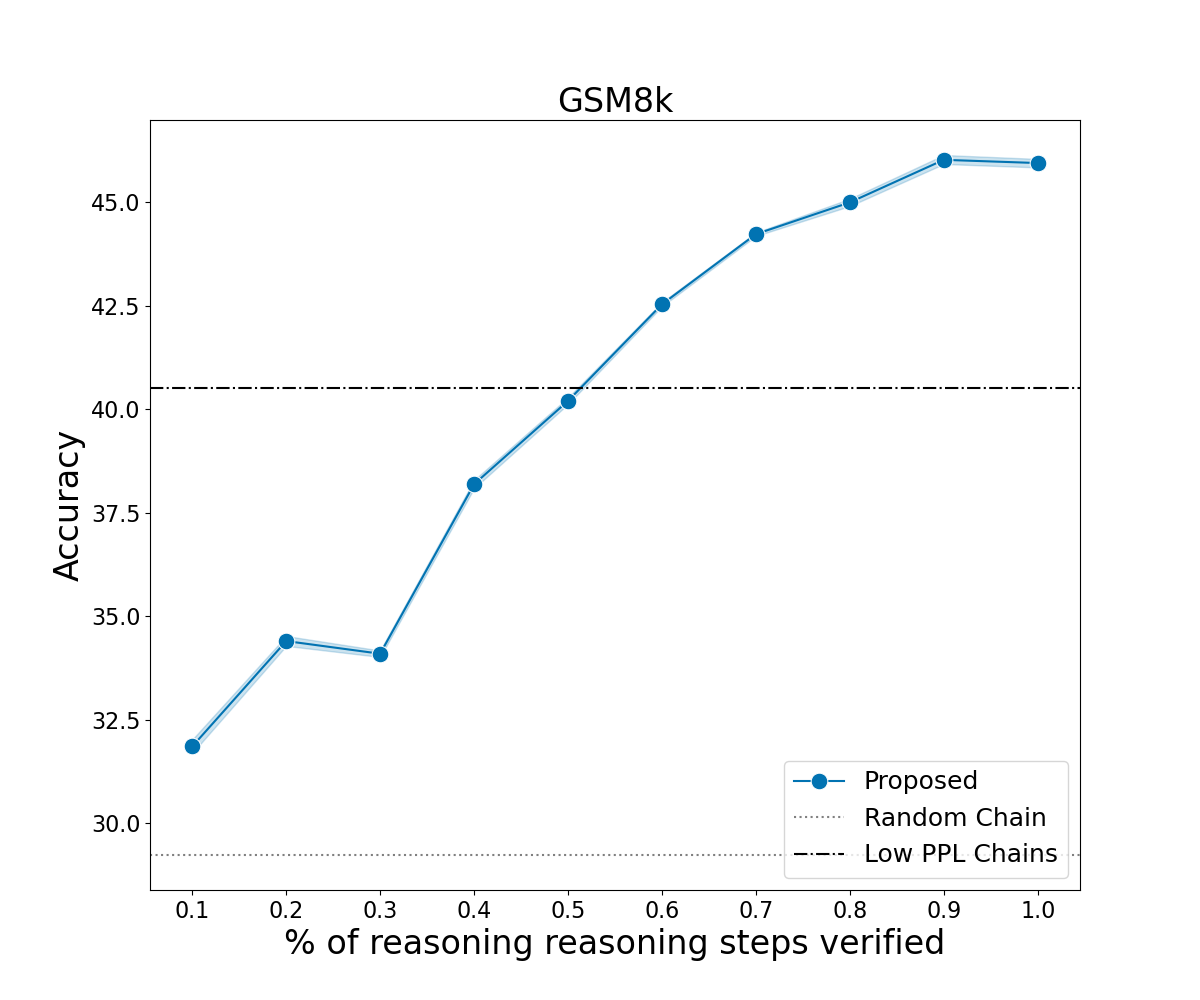}
        \caption{GSM8k}
        \label{l1}
    \end{subfigure}
    \hfill
    \begin{subfigure}[b]{0.31\textwidth}
        \includegraphics[width=1.1\columnwidth]{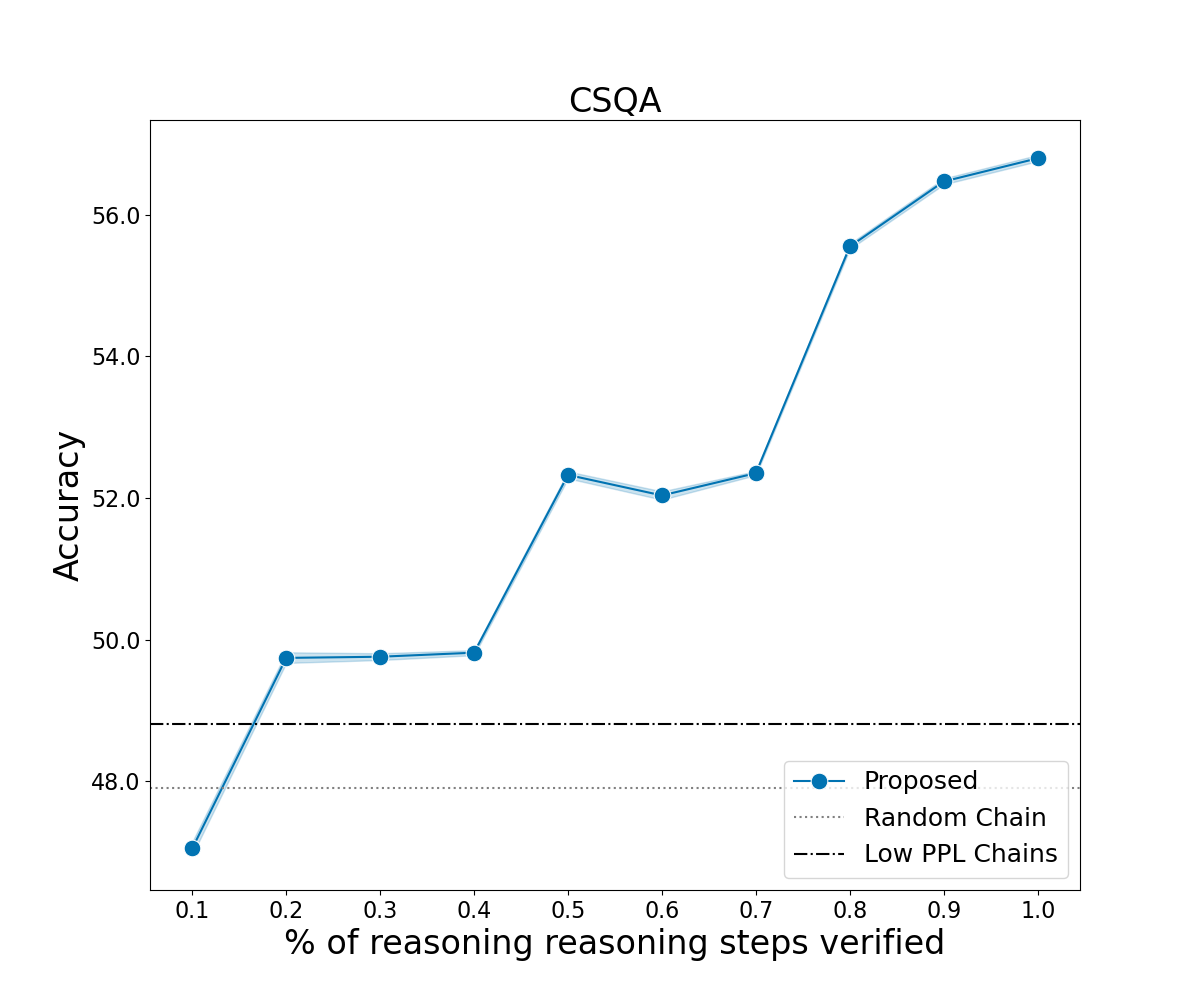}
        \caption{CSQA}
        \label{l1}
    \end{subfigure}
    \hfill
    \begin{subfigure}[b]{0.31\textwidth}
        \includegraphics[width=1.1\columnwidth]{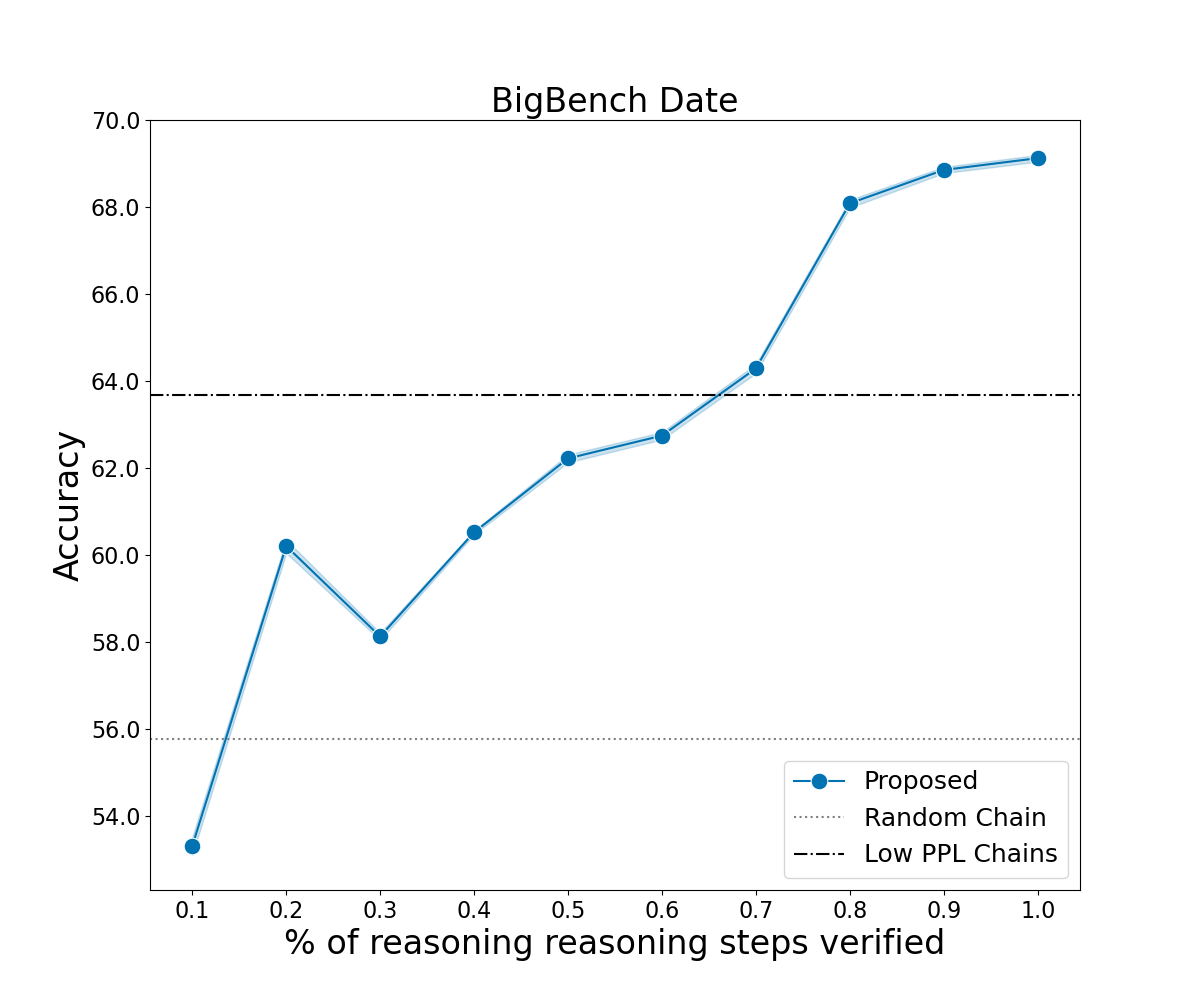}
        \caption{BigBench Date Understanding}
        \label{l1}
    \end{subfigure}
    \caption{Verifying only the first X\% steps of a given reasoning chain.  ($\uparrow$).}
    \label{fig:incomplete_rc_perc}
\end{figure*}

We assess the impact of using the verifiers for varying percentages of reasoning steps, denoted as X\% along the X-axis of our line plot in Figure~\ref{fig:incomplete_rc_perc}.
We remark that the final performance increases with the \% of steps verified and that verifying only the first 20\% of the steps is sufficient to increase the final performance beyond that of random chains.

Since knowing beforehand the total number of reasoning steps is unrealistic, we also experiment with only verifying a given number of the initial reasoning steps. Due to space limitations we include these results in Appendix~\ref{appendix:incomplete_rc}. Additionally, we include in Table~\ref{a:tab:incomplete_rc} the resulting performance when verifying between $0$ and \textit{All} reasoning steps over all datasets. In $7/9$ cases, the performance increases even when verifying only the first step. When verifying the first two steps, the final performance increases in all the cases.

\begin{table}[t]
    \begin{subtable}[b]{1\columnwidth}
        \resizebox{1.0\columnwidth}{!}{\begin{tabular}{rrrrrrr}
\toprule
\multicolumn{4}{c}{Verifiers} & \multicolumn{3}{c}{Math} \\
\cmidrule(lr){1-4}\cmidrule(lr){5-7}
P & R & M & C & AddSub & GSM8k & SVAMP \\
\midrule
\xmark & \xmark & \xmark & \xmark & 41.04$\pm$1.79 & 29.23$\pm$1.58 & 38.78$\pm$1.26 \\
\xmark & \xmark & \xmark & \cmark & 48.84$\pm$0.85 & 33.51$\pm$0.69 & 45.41$\pm$0.56 \\
\xmark & \xmark & \cmark & \xmark & 45.76$\pm$1.99 & 35.59$\pm$1.60 & 41.68$\pm$1.65 \\
\xmark & \cmark & \xmark & \xmark & 49.04$\pm$0.44 & 33.38$\pm$0.52 & 46.98$\pm$0.42 \\
\cmark & \xmark & \xmark & \xmark & 59.09$\pm$0.59 & 41.53$\pm$0.49 & 53.85$\pm$0.45 \\
\cmark & \cmark & \cmark & \cmark & \textbf{62.34$\pm$0.25} & \textbf{45.94$\pm$0.30} & \textbf{56.36$\pm$0.23} \\
\bottomrule
\end{tabular}
}
        \caption{Ablation over Math datasets.}
        \label{tab:ablation_sc_m}
    \end{subtable}
    \newline
    \vfill
    \begin{subtable}[b]{1\columnwidth}        
        \resizebox{1.0\columnwidth}{!}{\begin{tabular}{rrrrrrrrr}
\toprule
\multicolumn{3}{c}{Verifiers} & \multicolumn{1}{c}{Other} & \multicolumn{3}{c}{Commonsense} & \multicolumn{2}{c}{Symbolic} \\
\cmidrule(lr){1-3}\cmidrule(lr){4-4}\cmidrule(lr){5-7}\cmidrule(lr){8-9}
P & R & C & BigBench Date & CSQA2.0 & CSQA & Strategy & Coinflip & Last Letter\\
\midrule
\xmark & \xmark & \xmark & 55.77$\pm$3.06 & 58.75$\pm$1.68 & 47.91$\pm$1.22 & 56.03$\pm$0.99 & 58.67$\pm$2.04 & 15.68$\pm$1.51 \\
\xmark & \xmark & \cmark & 62.70$\pm$0.68 & 56.77$\pm$0.55 & 53.00$\pm$0.59 & 54.31$\pm$0.70 & 49.17$\pm$0.76 & 20.63$\pm$0.78 \\
\xmark & \cmark & \xmark & 62.57$\pm$0.45 & 61.03$\pm$0.21 & 52.62$\pm$0.29 & 57.31$\pm$0.26 & 53.63$\pm$0.51 & 15.79$\pm$0.30 \\
\cmark & \xmark & \xmark & 64.06$\pm$0.60 & 60.27$\pm$0.25 & 51.35$\pm$0.36 & \textbf{60.03$\pm$0.30} & \textbf{73.36$\pm$0.42} & \textbf{41.73$\pm$0.48} \\
\cmark & \cmark & \cmark & \textbf{69.12$\pm$0.21} & \textbf{62.16$\pm$0.22} & \textbf{56.79$\pm$0.12} & 57.21$\pm$0.17 & 64.02$\pm$0.21 & 41.64$\pm$0.44 \\
\bottomrule
\end{tabular}

}
        \caption{Ablation over Non-Math datasets.}
        \label{tab:ablation_sc_nm}
    \end{subtable}
    \caption{Ablation study on the effect of each verifier on the downstream tasks when selecting a \textit{single} reasoning chain. We differentiate between math and non-math datasets.}
    \label{fig:ablation_sc}
\end{table}

\subsection{Contributions of each Verifier}
In this experiment, we assess the contribution to the final performance of each of our verifiers: (1) Low Step \textbf{P}erplexity, (2) \textbf{R}elevance, (3) \textbf{M}athematical Accuracy (if applicable), (4) Logical \textbf{C}onsistency.
First, we observe that each individual verifier is meaningfully contributing towards the final solution. For example, for the math datasets (Table~\ref{tab:ablation_sc_m}), employing \textit{any} verifier improves the final performance, with improvements ranging from $2.90\%$ to $21.30\%$. All in all, using as little as a single verifier improves the final performance in over 89\% of the cases.\footnote{35/39} 
Secondly, we remark that combining all the verifiers gives further improvements, beyond those obtained by using a single verifier, suggesting that each verifier is adding meaningful and non-overlapping information. We note that there is a notable exception to this trend, where for the Symbolic Reasoning tasks (and for the Strategy dataset), a distinct combination of verifiers (i.e. only Perplexity) attains a better score than using all the verifiers. We provide more comprehensive results covering a wider range of verifier combinations in Appendix~\ref{appendix:ablation}.

\subsection{Human Evaluation}
\label{s:he}
While the proposed verifiers meaningfully contribute to the final performance on the reasoning downstream tasks, we perform a human evaluation study to assess: (1) how well they correlate with human judgment, (2) how reliably concepts such as logical consistency or relevance can be evaluated by humans, and (3) the percentage of errors not captured by the proposed principles. 
We randomly sample candidate reasoning chains for 3 datasests: GSM8k, CSQA2.0, and Coinflip. We then distribute these to 8 human annotators and ask them to annotate each reasoning chain according to the following four criteria: (i) Relevance: is this step helpful in reaching the final solution, (ii) Mathematical Accuracy: are the mathematical calculations (if any) correct, (iii) Consistency: is the current step consistent with the previous steps, and (iv) Overall Correctness. We include inter-annotator agreement scores in the Appendix~\ref{appendix:human_eval}. 

\begin{figure}[t]
   \begin{center}
   \includegraphics[width=1.0\columnwidth]{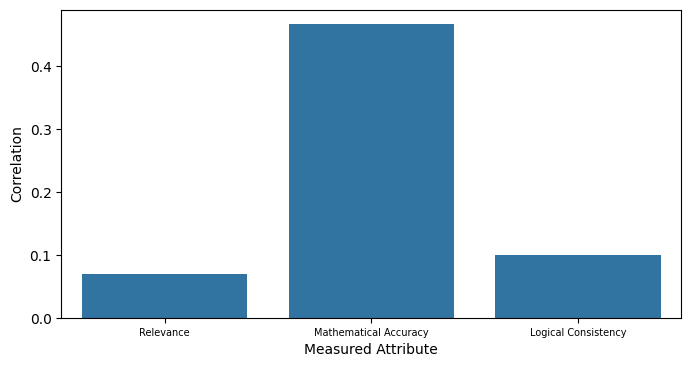}
   \end{center}
   \caption{
   Correlation between the scores of our proposed verifiers and the assessment of human annotators over the three reasoning principles explored in this work. All correlations have a p-value of less than $0.0001$.
   }
   \label{fig:correlation_verifiers}
\end{figure}

We compute pearson correlation scores (the correlation scores are in $[-1, 1])$ between human assessments and the scores proposed by our verifiers and summarize the results in Figure~\ref{fig:correlation_verifiers}. We draw the following conclusions.

First, we remark that each verifier exhibits a significant ($\text{p-value} < 0.0001$) and positive correlation (although small for Relevance and Logical Consistency) with human judgment. 
Given the empirical improvements we observed when using the verifiers and the low correlations between the verifiers and the humans, one might wonder whether the improvements we have empirically observed are reasonable.
To address this concern, we conducted additional experiments on artificially generated data, where we had precise control over the correlation values. We provided additional details in Appendix~\ref{a:s:performance_given_correlation}. This experiment showed that the improvements we have observed are expected and even small correlations can statistically differentiate better reasoning chains from worse ones on average.
We leave the exploration of verifiers that correlate more strongly with human judgment to future work.

Second, we observed a large variance in the inter-annotator agreement score, which we hypothesize to come from different reasoning styles between different humans. 
This variation is reminiscent of findings in toxicity research, where annotator backgrounds influence their judgments \cite{Goyal2022IsYT}.\footnote{We first observed this in the pilot study, after which we improved the clarity of the guidelines.} We show all the inter-annotator agreements over each of the four attributes in Appendix~\ref{appendix:human_eval}.

Lastly, we found that less than 2\% of the errors marked by the annotators are not captured by one of the three principles explored.

\section{Conclusion}
In this study, we explore the applicability of a general-purpose verification procedure that can be readily applied to a wide range of reasoning tasks. This verification procedure consists of task- and dataset-agnostic verifiers designed to operate at the reasoning step-level and is inspired from fundamental principles that underlie sound reasoning: (1) Relevance (2) Mathematical Accuracy and (3) Logical consistency. On top of these, we leverage the perplexity of the reasoning step to steer the LLM towards high-quality solutions.

We evaluated our proposed approach across four distinct reasoning tasks, spanning nine datasets. Our results consistently demonstrate that the adoption of our proposed verifiers leads to notable performance improvements when compared to randomly sampled reasoning chains. Most notably, our approach outperforms the lowest perplexity reasoning chain in over 6 out of the 9 datasets we tested, indicating that the proposed verifiers provide additional information beyond what is captured by the perplexity measure of the reasoning chain. 

Overall, we found that large language models are capable of finding their own mistakes, albeit this capability is currently noisy. We leave the exploration of better and more cost-efficient verifiers to future work.

\section*{Acknowledgments}
We would like to thank Aaron Doliana, Anya Stallman, Claudia Zaghi, Elizabeth Okada, Elvira Magomedova, Emily Sheffield, Harrison Freck, I-Hsuan Chen, Mario Piergallini, Matt Ferguson, Paul Goeden for their help with the Chain of Thought annotations. We would like to thank Dan Roth and Ramesh Nallapati for fruitful discussions.

\section*{Limitations}

While our proposed framework is flexible and admits different implementations for the verifiers, in this work we implemented each verifier with a prompt-based LLM approach. This type of implementation can increase the energy consumption of the deployed system, leading to a performance--energy-consumption trade-off.  
Secondly, employing step-by-step verifiers increases the computational time needed by the system to produce an output. 
This trade-off must be analyzed on a case-by-case basis and compared to other alternatives, such as self-consistency. 
Different from self-consistency, our proposed approach aims to improve the correctness of each step in a step-by-step solution.

While our evaluation spanned multiple reasoning tasks and datasets, it was limited to tasks in the English language only.

During our human evaluation study, we have observed low positive (but significant) correlations. 

Lastly, we only used Falcon-40B-Instruct in our experiments, as it was one of the largest and most capable open-source\footnote{Open source according to https://opensource.org/} model at the time of the experiments and did not evaluate with more powerful but closed-source models such as GPT-4.

\section*{Ethics Statement}
Our proposed approach utilizes large language models, which are known to be biased and to hallucinate. In this work, we do not pre-train nor fine-tune any large-scale models. Instead, we use already pre-trained open-source models and prompting.


\bibliography{custom}

\appendix
\section{Experimental Settings}

\subsection{Datasets}
\label{appendix:datasets}
We include in Table~\ref{a:tab:dataset} an input/output example for each dataset used.
We also experimented with the Object Tracking problem from BigBench, but we removed it because for all experimental settings, encompassing both baseline and proposed methods, the performance consistently fell below the chance-level threshold. 
For Last Letter Concatenation, we use only $2$ words instead of $4$, as we empirically observed that Falcon-40B is not able to tackle the problem when there are four words.
We use a temperature of $0.7$ for all our experiments.

\subsection{Hardware}
We ran the experiments on machines with 8 A10G 24GB GPUs (AWS \texttt{g5.48xlarge}). In total, we used approximately 10 weeks worth of \texttt{g5.48xlarge} time.

\begin{table*}[]
\begin{tabular}{p{0.2\textwidth}p{0.15\textwidth}p{0.5\textwidth}p{0.1\textwidth}}
\toprule
\textbf{Reasoning Task}               & \textbf{Dataset Name}                & \textbf{Input}                                                                                                                                                                                         & \textbf{Expected Output}       \\ \midrule
Other                        & BigBench Date Understanding & Yesterday was April 30, 2021. What is the date today in MM/DD/YYYY? Choices: (A) 05/01/2021, (B) 02/23/2021, (C) 03/11/2021, (D) 05/09/2021, (E) 06/12/2021, (F) 04/29/2021                   & (A) \\ \midrule
\multirow{2}{*}{Commonsense} & CSQA 2.0                    & a pupil can be either a student or part of an eye                                                                                                                                             & yes        \\
                             & Strategy                    & Is it common to see frost during some college commencements?                                                                                                                                  & yes        \\ \midrule
\multirow{2}{*}{Symbolic}    & Coinflip                    & A coin is heads up. Whitney flips the coin. Erika does not flip the coin. Tj does not flip the coin. Benito flips the coin. Is the coin still heads up? Note that "flip" here means "reverse" & yes        \\
                             & Last Letter (2)             & Take the last letters of each words in "Whitney Benito" and concatenate them.                                                                                                                 & yo         \\ \midrule
\multirow{3}{*}{Math}        & GSM8k                       & Natalia sold clips to 48 of her friends in April, and then she sold half as many clips in May. How many clips did Natalia sell altogether in April and May?                                   & 72         \\
                             & SVAMP                       & Bryan took a look at his books as well. If he has 34 books distributed equally in 2 bookshelves. How many books are there in each bookshelf?                                                  & 17         \\
                             & AddSub                      & Joan found 70 seashells on the beach. she gave Sam some of her seashells. She has 27 seashell. How many seashells did she give to Sam ?                                                       & 43         \\ \bottomrule
\end{tabular}
\caption{An example of input and expected output for each of the datasets we experiment with.}
\label{a:tab:dataset}
\end{table*}

\section{Verifiers}
\label{appendix:verifiers}
We include the prompts we used for each verifier.
\subsection{Relevance Prompt}
\begin{quote}
You are a helpful assistant that is good at evaluating reasoning chains in order to solve logic problems.\\
You are given a logic problem and a draft solution with numbered steps that we need to complete. Evaluate the draft solution by determining whether it adds relevant information that helps to solve the problem. If it is relevant answer by 'yes, the solution is relevant', otherwise say 'no' and explain which steps failed. \\
\#\#\# Problem: \{problem statement\} \\
\#\#\# Draft solution: \{previous steps\} \\
\#\#\# Draft step: \{current step\} \\
\#\#\# Your evaluation of the draft step:
\end{quote}

\subsection{Mathematical Accuracy Prompt}
For brevity, we have included a single example from the set of 20 in-context examples. The in-context examples are a mix of inputs with mathematical calculations and without. Furthermore, within the set of examples involving mathematical calculations, both incorrect and correct calculations are included for comprehensive coverage.
\begin{quote}
Instruction:\\
The task is to extract the mathematical calculations appearing below and return the result in JSON format. Please do not perform any additional calculations and do not introduce any number or numerical expression that does not appear in the original input text. If there is no explicit calculation performed, do not return anything.\\\\
Input:\\
Therefore, he has \$87-\$32=<<87-32=40>>\$40 left\\
\\
Output:\\
```json\\
\{[\{"lhs": "87-32", "op": "=", "rhs": "40"\}]\}\\
```\\
<..>
Input:\\
\{input\}\\

Output:\\
```json\\

\end{quote}

\subsection{Logical Consistency Prompt}
\begin{quote}
"""You are a smart, critical, and logical teacher assistant. You are critically reading a student's answer line by line and verifying each line for any contradictions in the student's argument. More information below.\\
\\
Previous Steps:\\
\{previous steps\}\\
\\
Last Step:\\
\{current step\}\\
\\
Instruction:\\
Given the information present in the Last Step and in the Previous Steps, please check if the conclusion present in the Last Step is contradicting any information from the Previous Steps.\\
\\
Feedback:\\
Based on the Last Step, which is "\{current step\}", and on the Previous Steps, we can conclude that the Last Step is"""
\end{quote}

\section{Self-Consistency}
\label{appendix:self_consistency}
We include here the plots for all the datasets for the Self-Consistency experiment performed in Section~\ref{ss:self_consistency}

\begin{figure*}[!t]
    \begin{subfigure}[b]{0.31\textwidth}
        \centering
        \includegraphics[width=0.8\textwidth]{figures/2309/BigBenchDate_sc.png}
        \caption{\footnotesize{BigBench Date Understanding}}
        \label{a:sc:d1}
    \end{subfigure}
    \hfill
    \begin{subfigure}[b]{0.31\textwidth}
        \centering
        \includegraphics[width=0.8\textwidth]{figures/2309/CSQA_sc.png}
        \caption{\footnotesize{CSQA}}
        \label{a:sc:d2}
    \end{subfigure}
    \hfill
    \begin{subfigure}[b]{0.31\textwidth}
        \centering
        \includegraphics[width=0.8\textwidth]{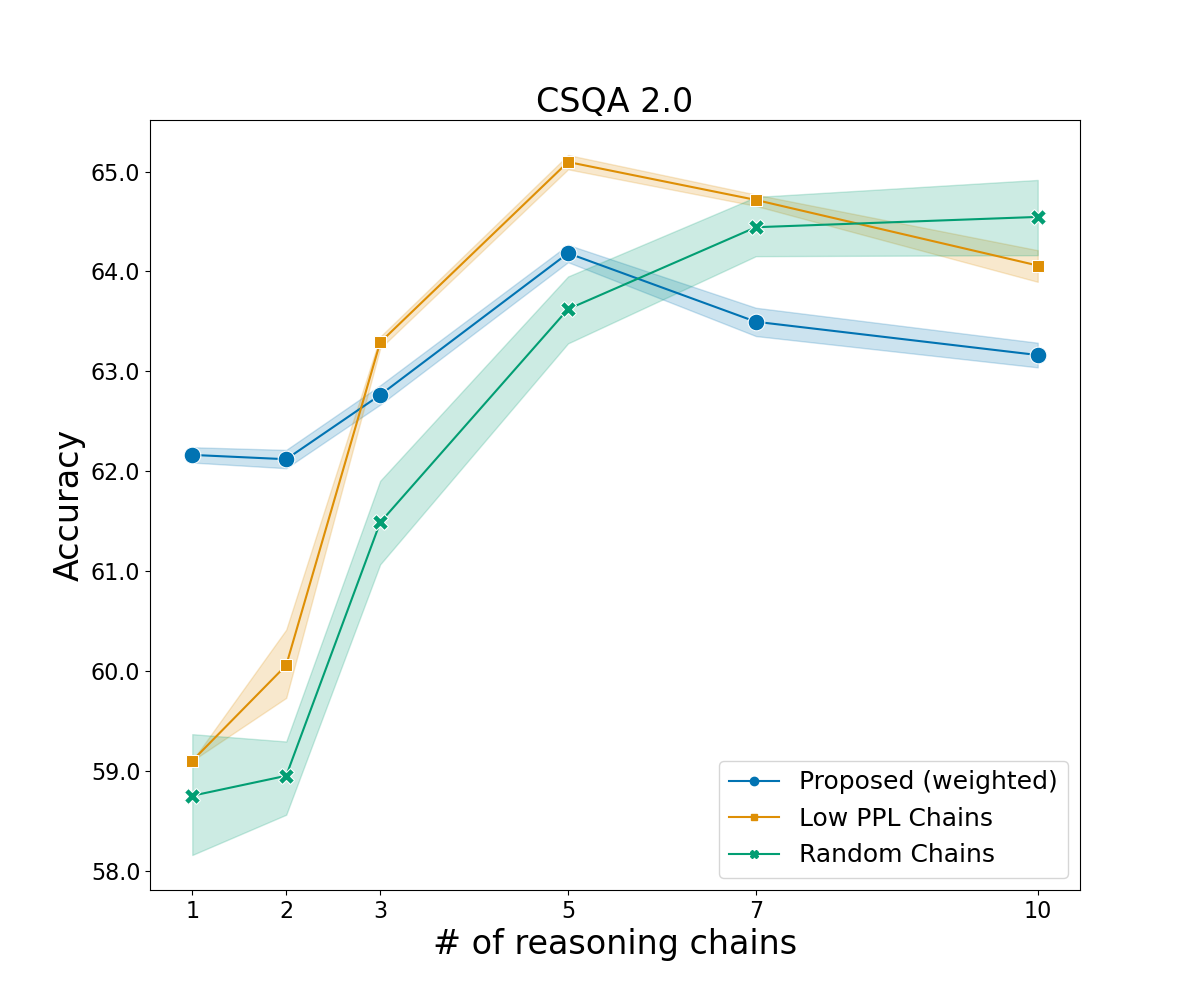}
        \caption{\footnotesize{CSQA 2.0}}
        \label{a:sc:d3}
    \end{subfigure}
    \hfill
    \\
    \begin{subfigure}[b]{0.31\textwidth}
        \centering
        \includegraphics[width=0.8\textwidth]{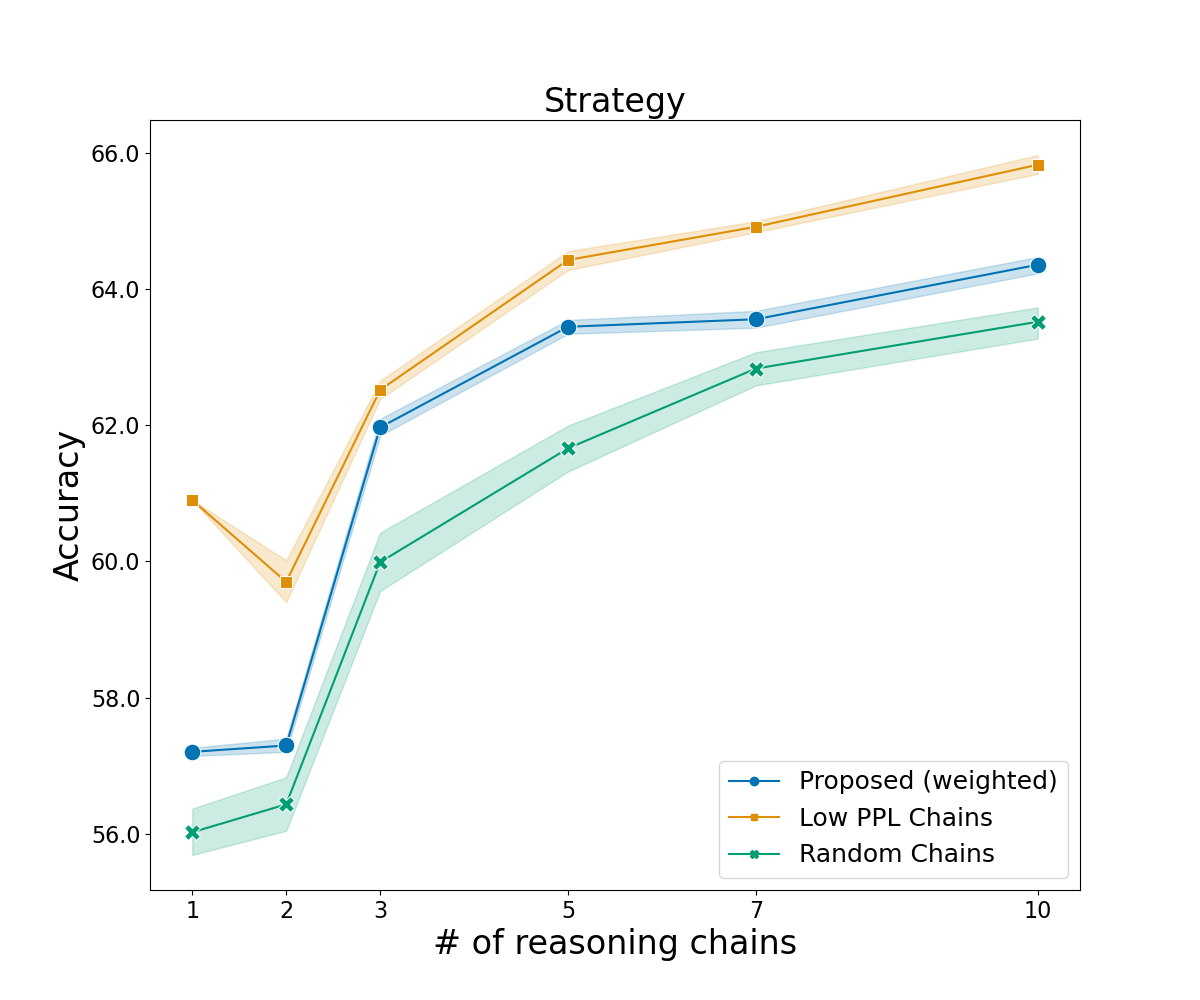}
        \caption{\footnotesize{Strategy}}
        \label{a:sc:d4}
    \end{subfigure}
    \hfill
    \begin{subfigure}[b]{0.31\textwidth}
        \centering
        \includegraphics[width=0.8\textwidth]{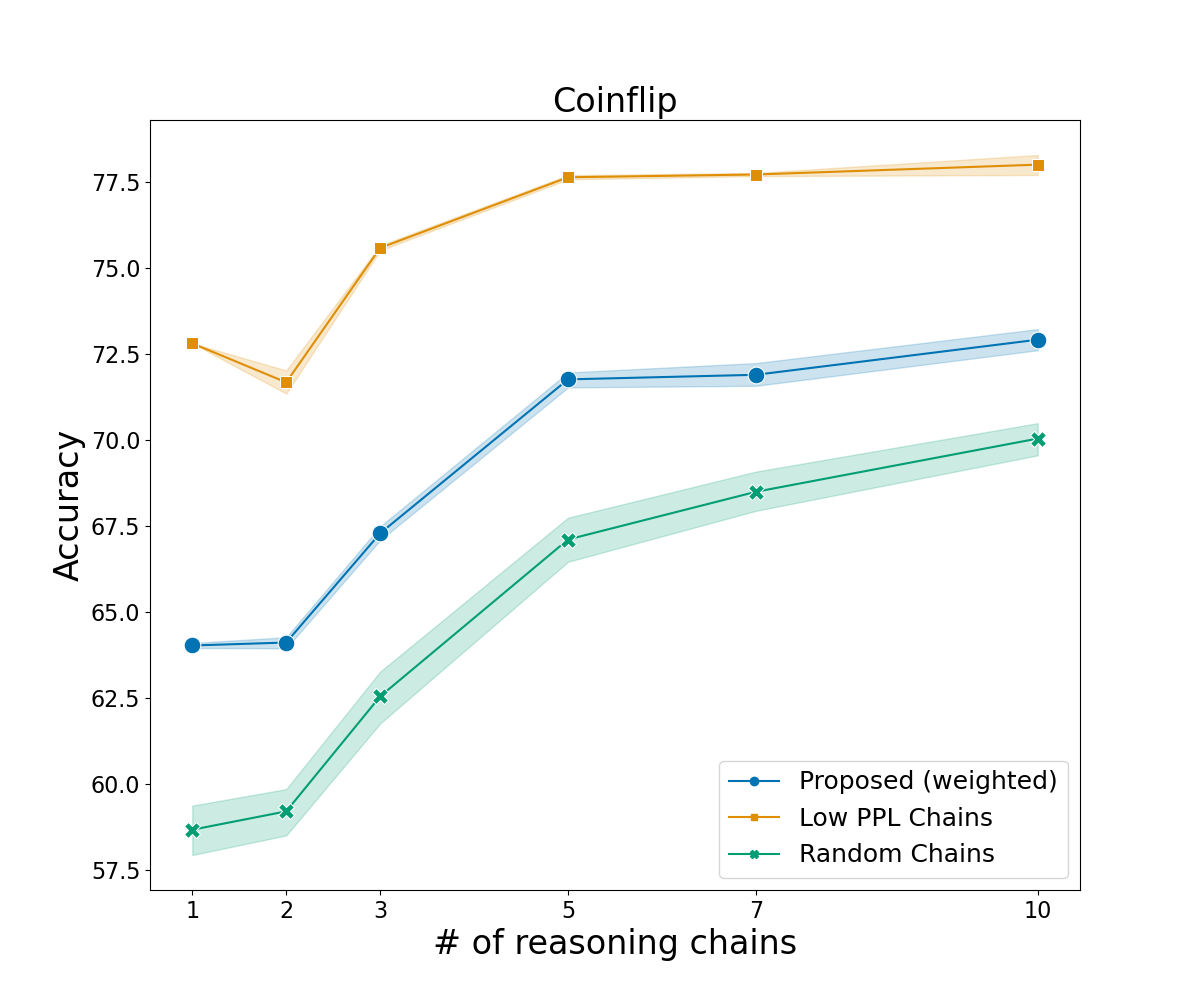}
        \caption{\footnotesize{Coinflip}}
        \label{a:sc:d5}
    \end{subfigure}
    \hfill
    \begin{subfigure}[b]{0.31\textwidth}
        \centering
        \includegraphics[width=0.8\textwidth]{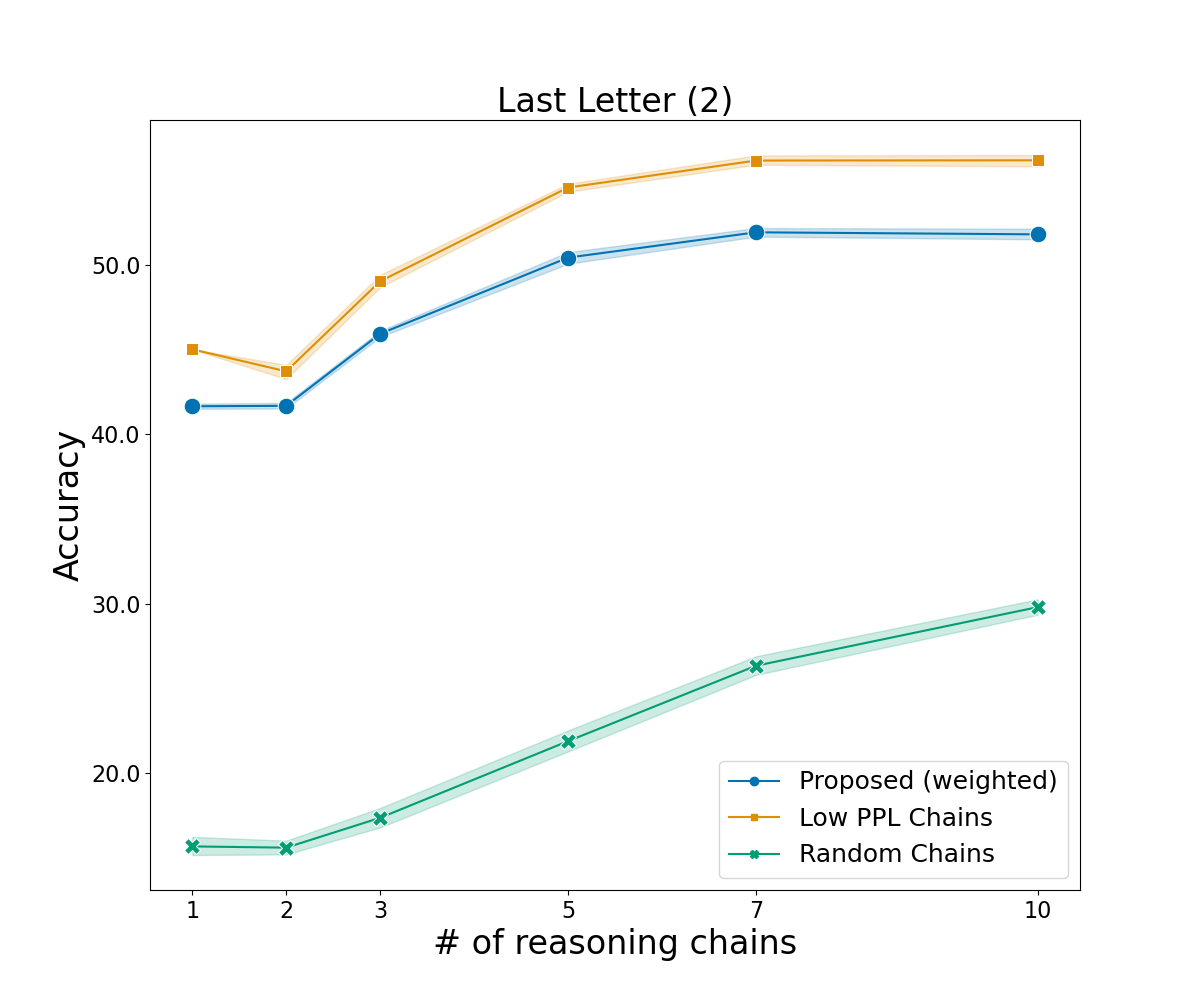}
        \caption{\footnotesize{Last Letter (2)}}
        \label{a:sc:d6}
    \end{subfigure}
    \hfill
    \\
    \begin{subfigure}[b]{0.31\textwidth}
        \centering
        \includegraphics[width=0.8\textwidth]{figures/2309/GSM8k_sc.png}
        \caption{\footnotesize{GSM8k}}
        \label{a:sc:d7}
    \end{subfigure}
    \hfill
    \begin{subfigure}[b]{0.31\textwidth}
        \centering
        \includegraphics[width=0.8\textwidth]{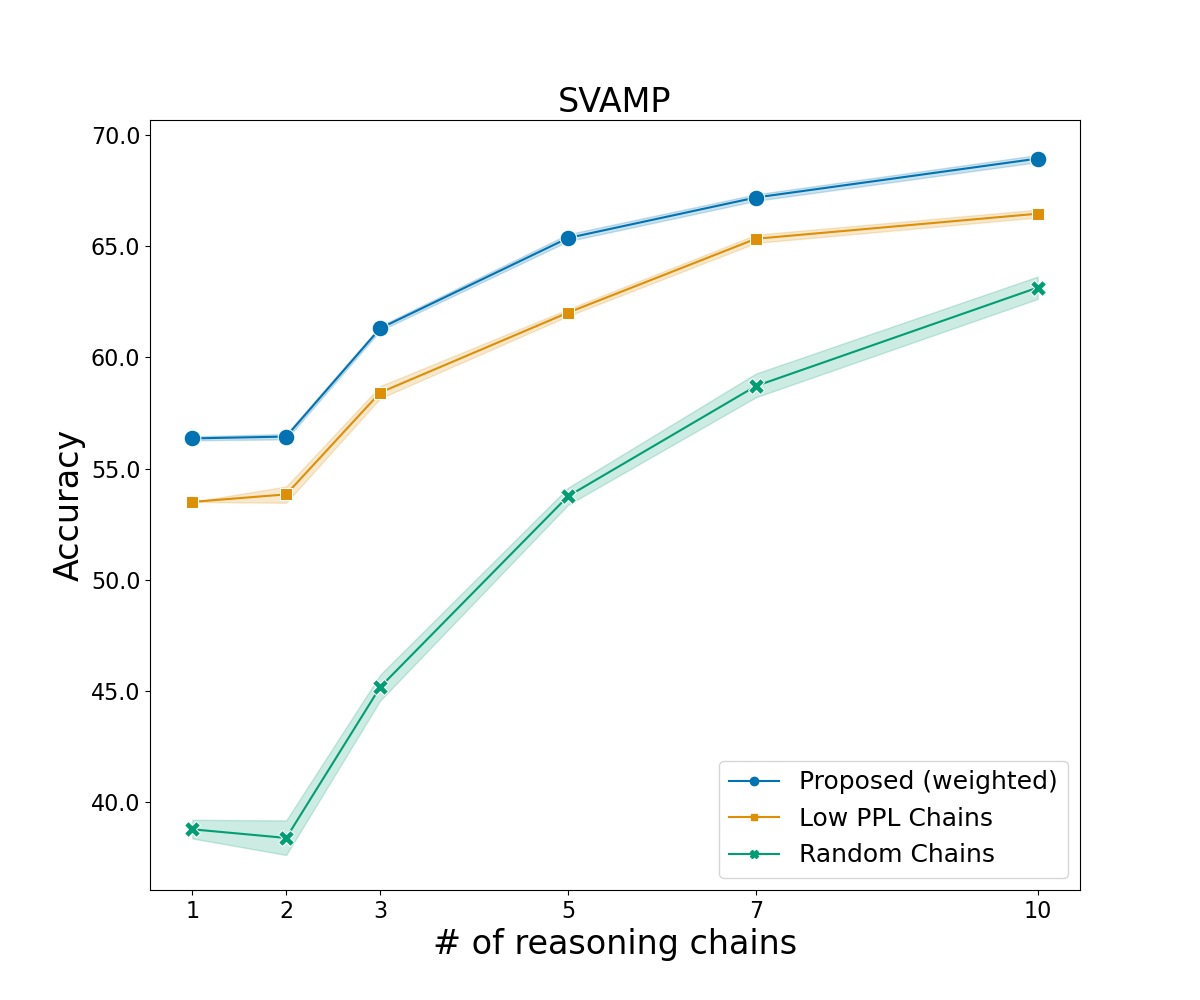}
        \caption{\footnotesize{SVAMP}}
        \label{a:sc:d8}
    \end{subfigure}
    \hfill
    \begin{subfigure}[b]{0.31\textwidth}
        \centering
        \includegraphics[width=0.8\textwidth]{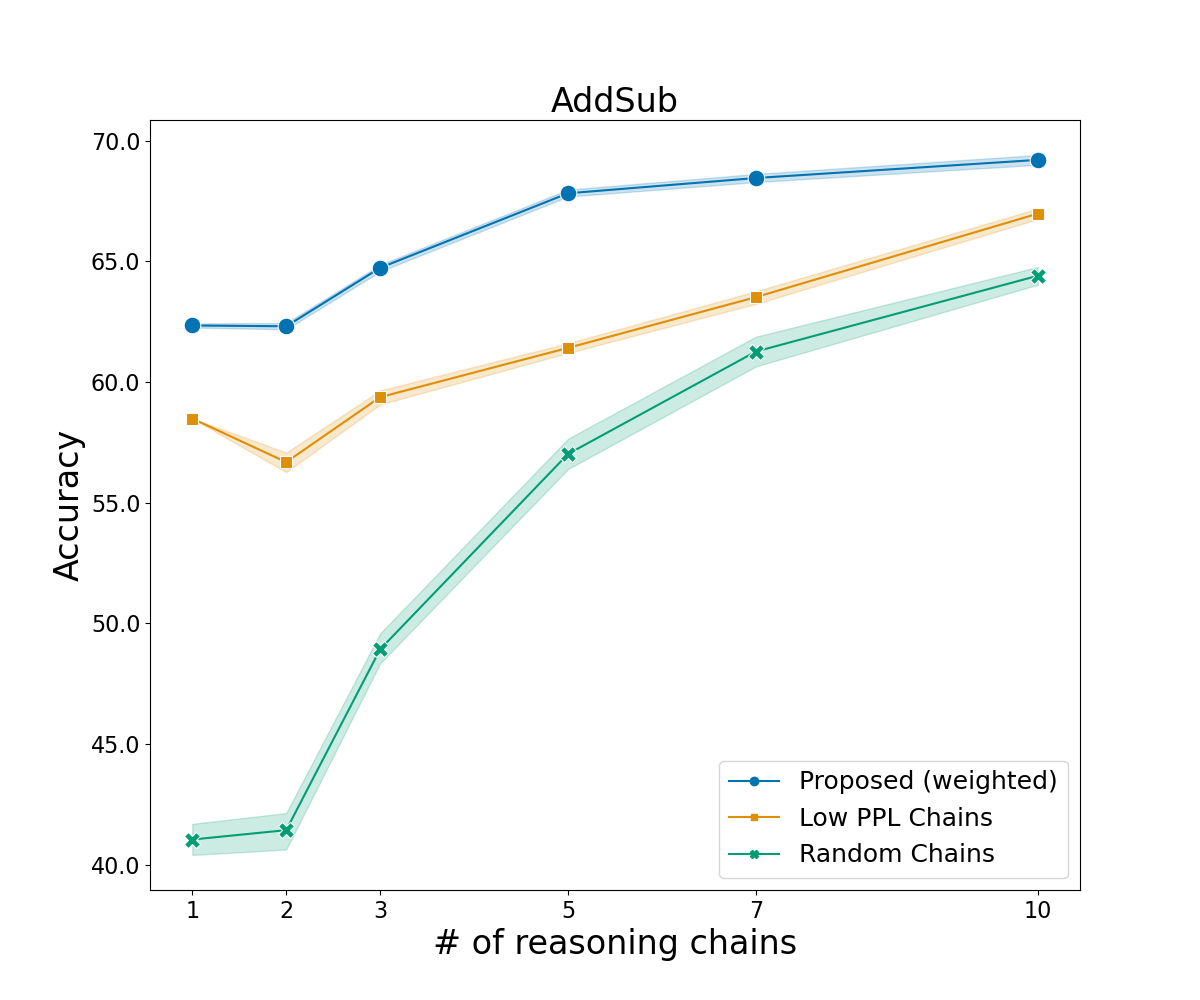}
        \caption{\footnotesize{AddSub}}
        \label{a:sc:d9}
    \end{subfigure}
    \hfill
    \caption{Self-consistency, sampling between 1-10 reasoning paths ($\uparrow$)}
    \label{appendix:fig:sc_train}    
\end{figure*}

\section{Self-Consistency: Weighted Voting vs Majority Voting}
\label{appendix:sc_voting}
\begin{figure*}[!t]
    \begin{subfigure}[b]{0.31\textwidth}
        \centering
        \includegraphics[width=0.8\textwidth]{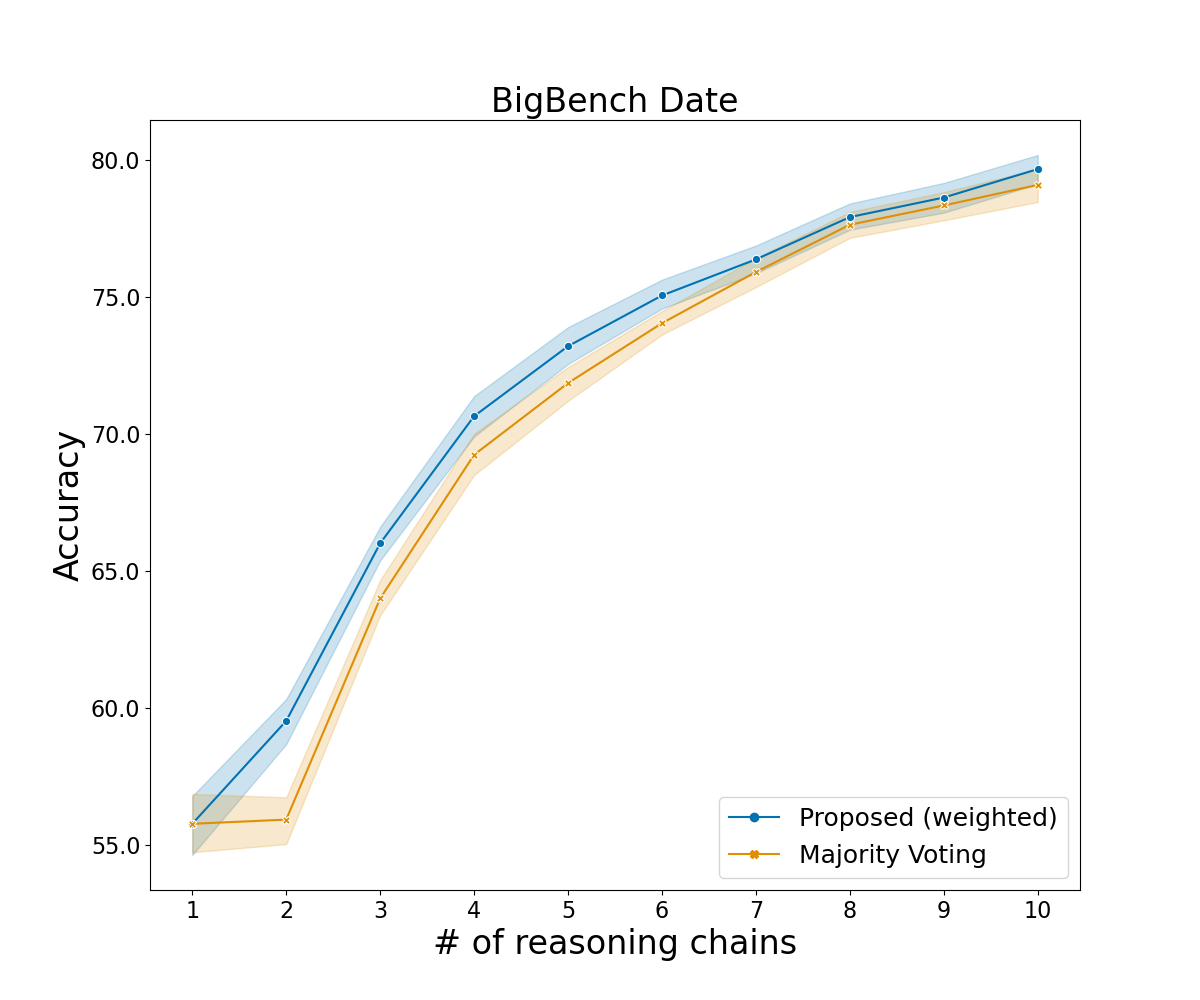}
        \caption{\footnotesize{BigBench Date Understanding}}
        \label{a:sc_wv_vs_ms:d1}
    \end{subfigure}
    \hfill
    \begin{subfigure}[b]{0.31\textwidth}
        \centering
        \includegraphics[width=0.8\textwidth]{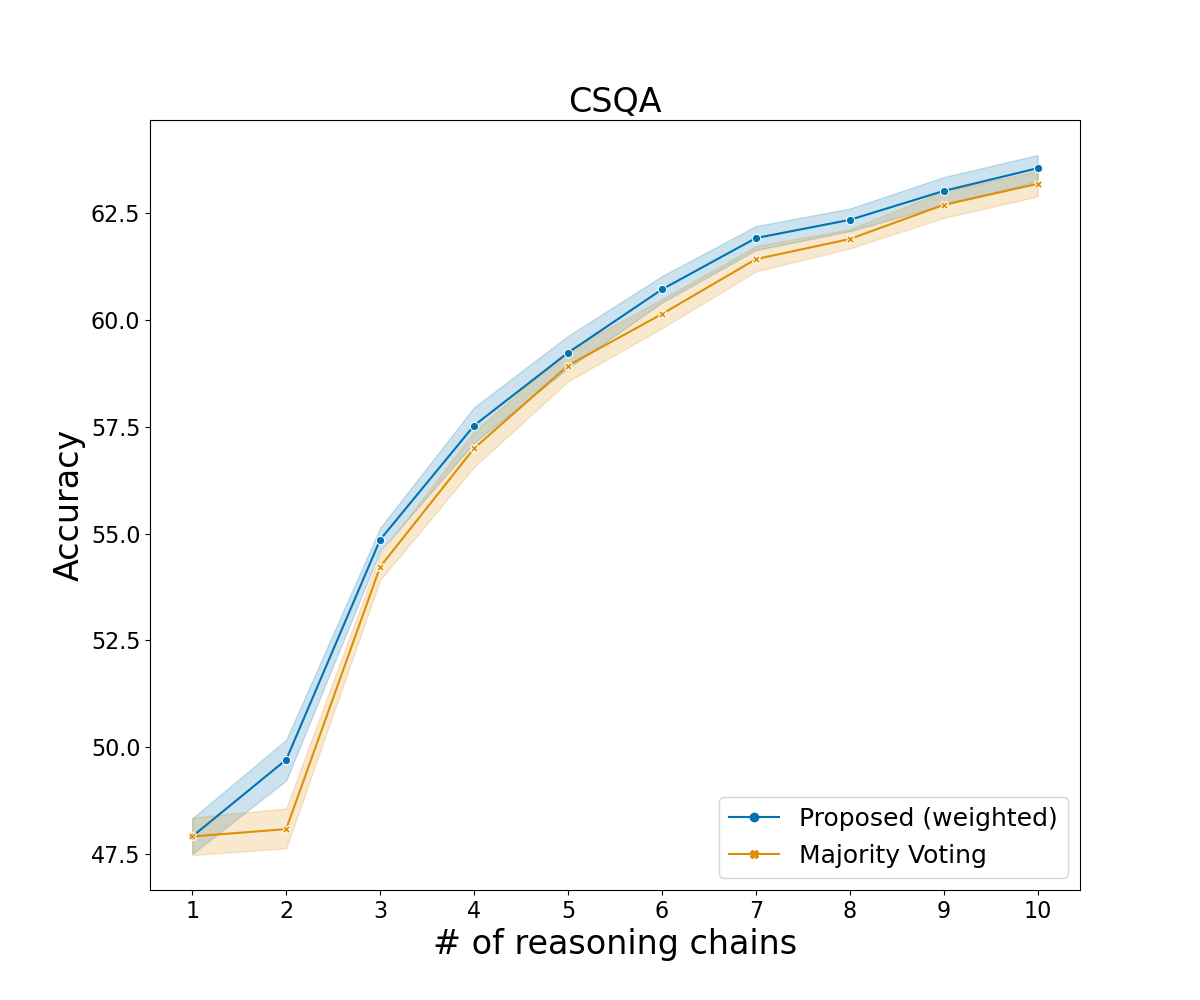}
        \caption{\footnotesize{CSQA}}
        \label{a:sc_wv_vs_ms:d2}
    \end{subfigure}
    \hfill
    \begin{subfigure}[b]{0.31\textwidth}
        \centering
        \includegraphics[width=0.8\textwidth]{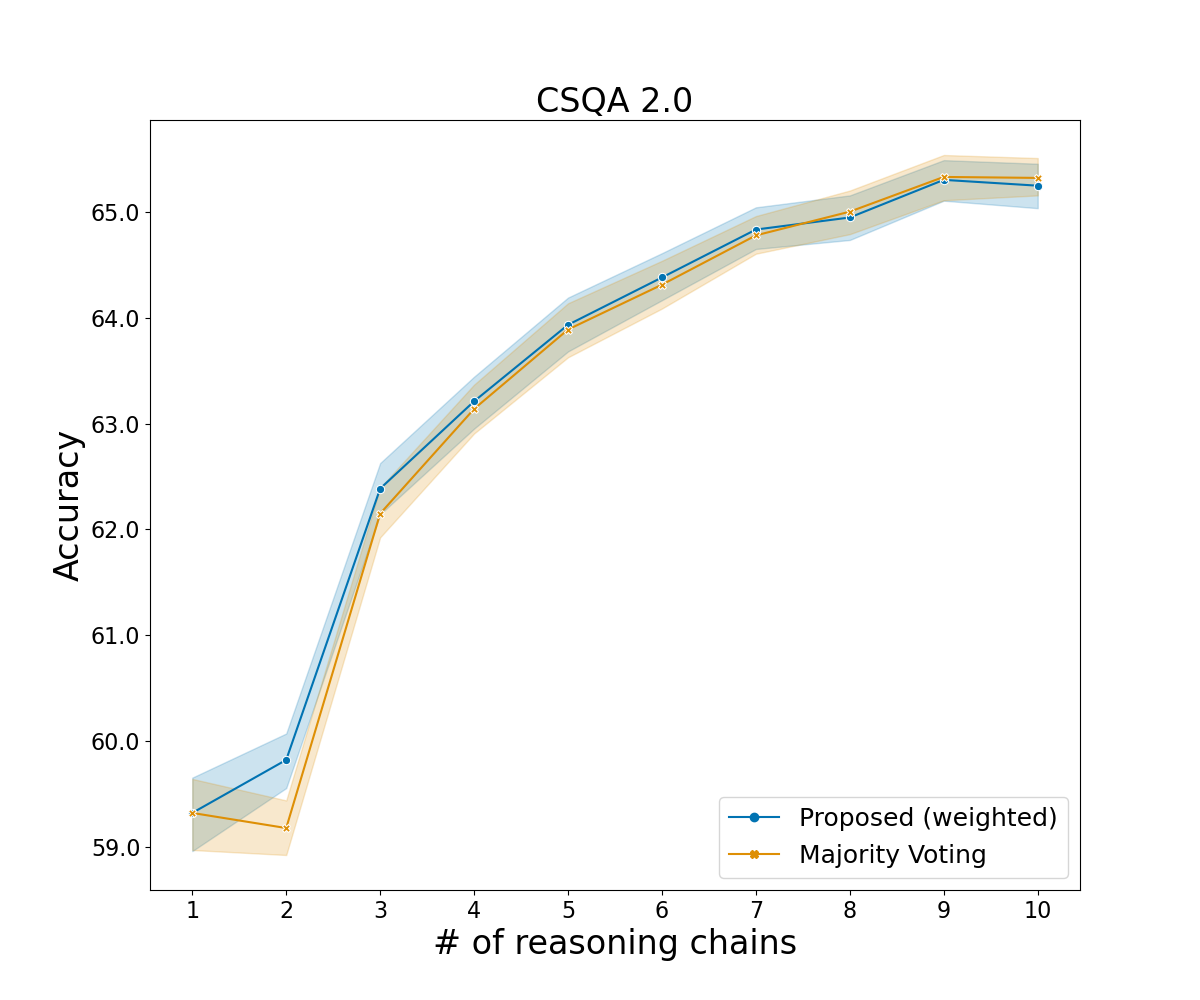}
        \caption{\footnotesize{CSQA 2.0}}
        \label{a:sc_wv_vs_ms:d3}
    \end{subfigure}
    \hfill
    \\
    \begin{subfigure}[b]{0.31\textwidth}
        \centering
        \includegraphics[width=0.8\textwidth]{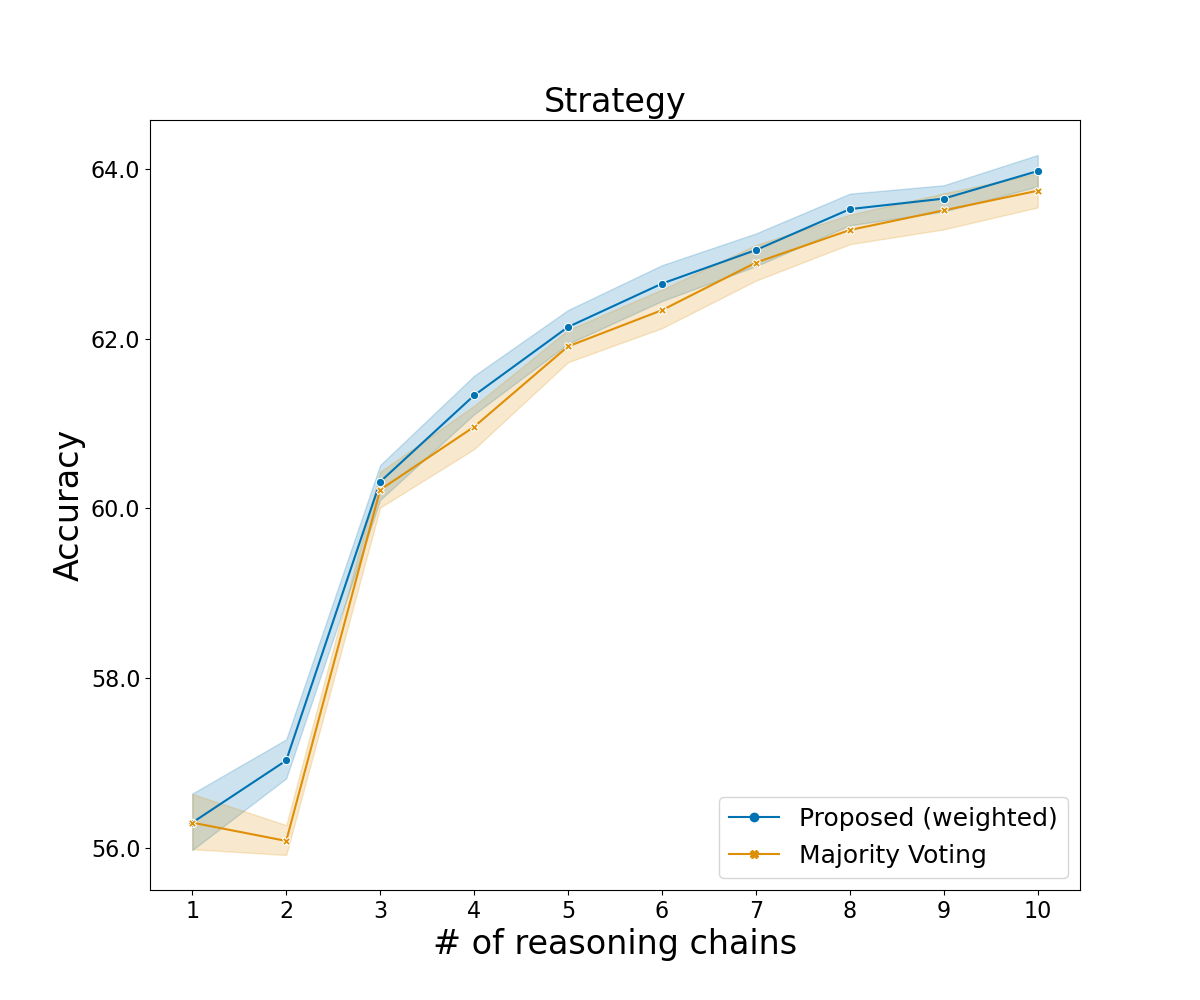}
        \caption{\footnotesize{Strategy}}
        \label{a:sc_wv_vs_ms:d4}
    \end{subfigure}
    \hfill
    \begin{subfigure}[b]{0.31\textwidth}
        \centering
        \includegraphics[width=0.8\textwidth]{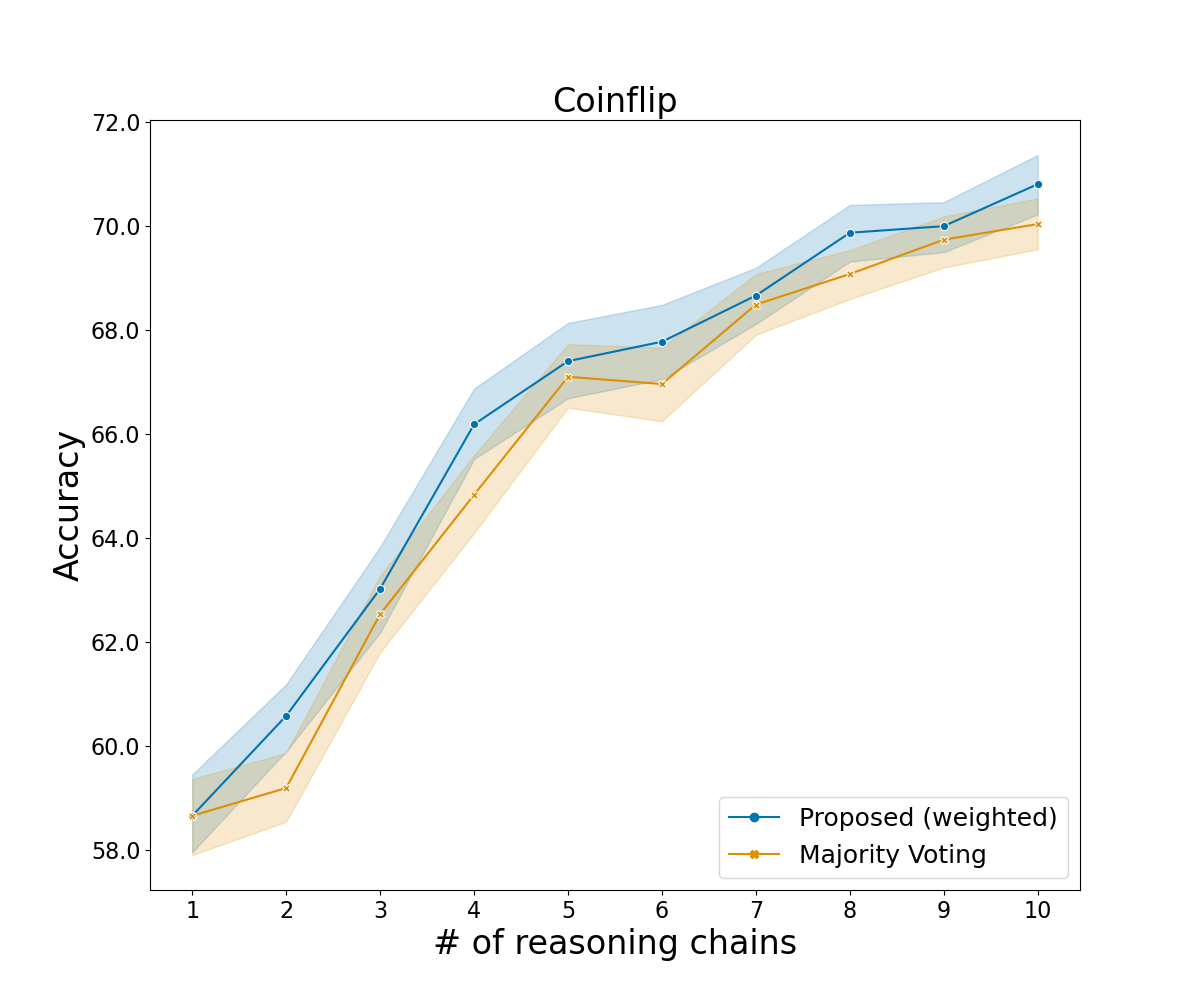}
        \caption{\footnotesize{Coinflip}}
        \label{a:sc_wv_vs_ms:d5}
    \end{subfigure}
    \hfill
    \begin{subfigure}[b]{0.31\textwidth}
        \centering
        \includegraphics[width=0.8\textwidth]{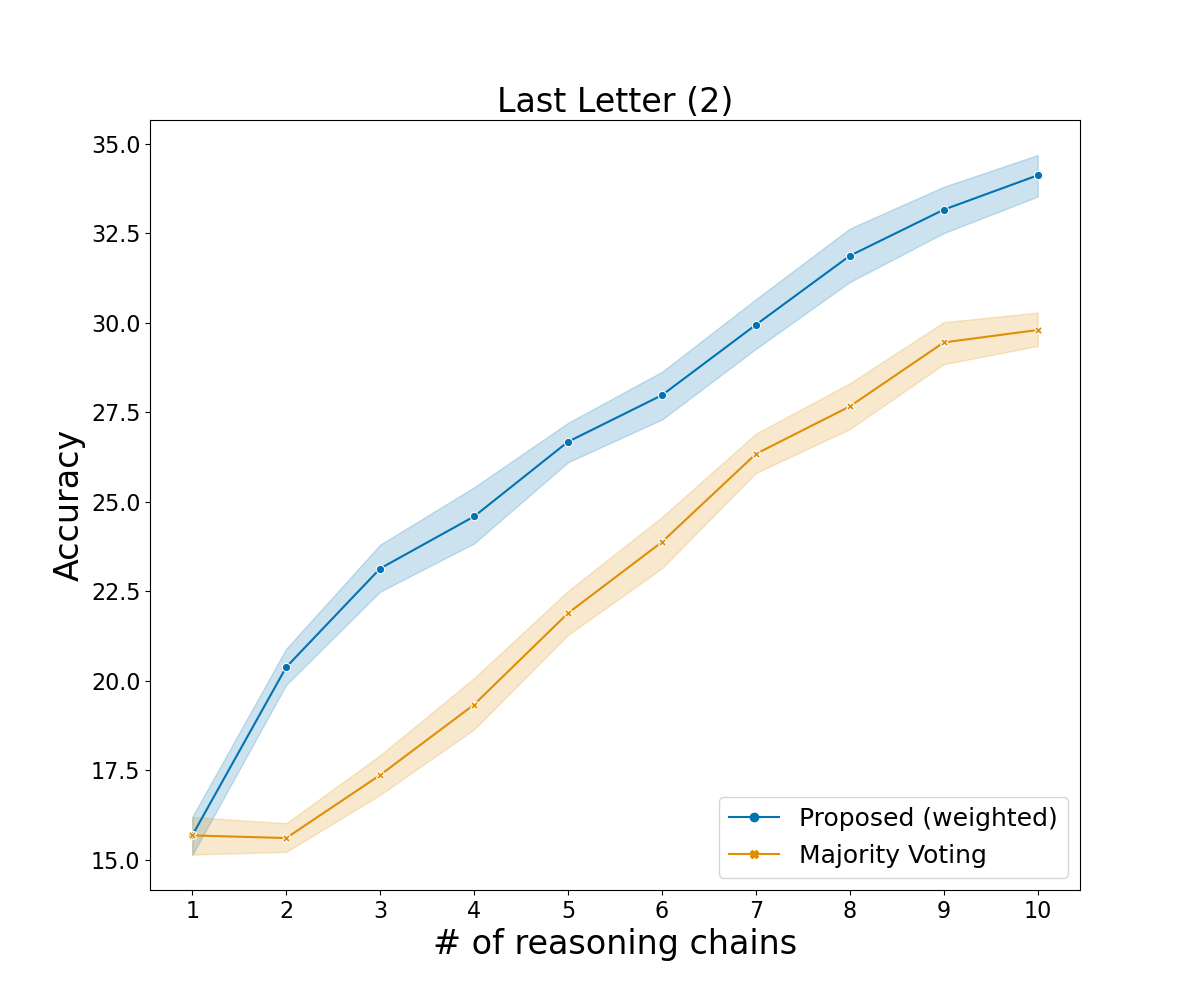}
        \caption{\footnotesize{Last Letter (2)}}
        \label{a:sc_wv_vs_ms:d6}
    \end{subfigure}
    \hfill
    \\
    \begin{subfigure}[b]{0.31\textwidth}
        \centering
        \includegraphics[width=0.8\textwidth]{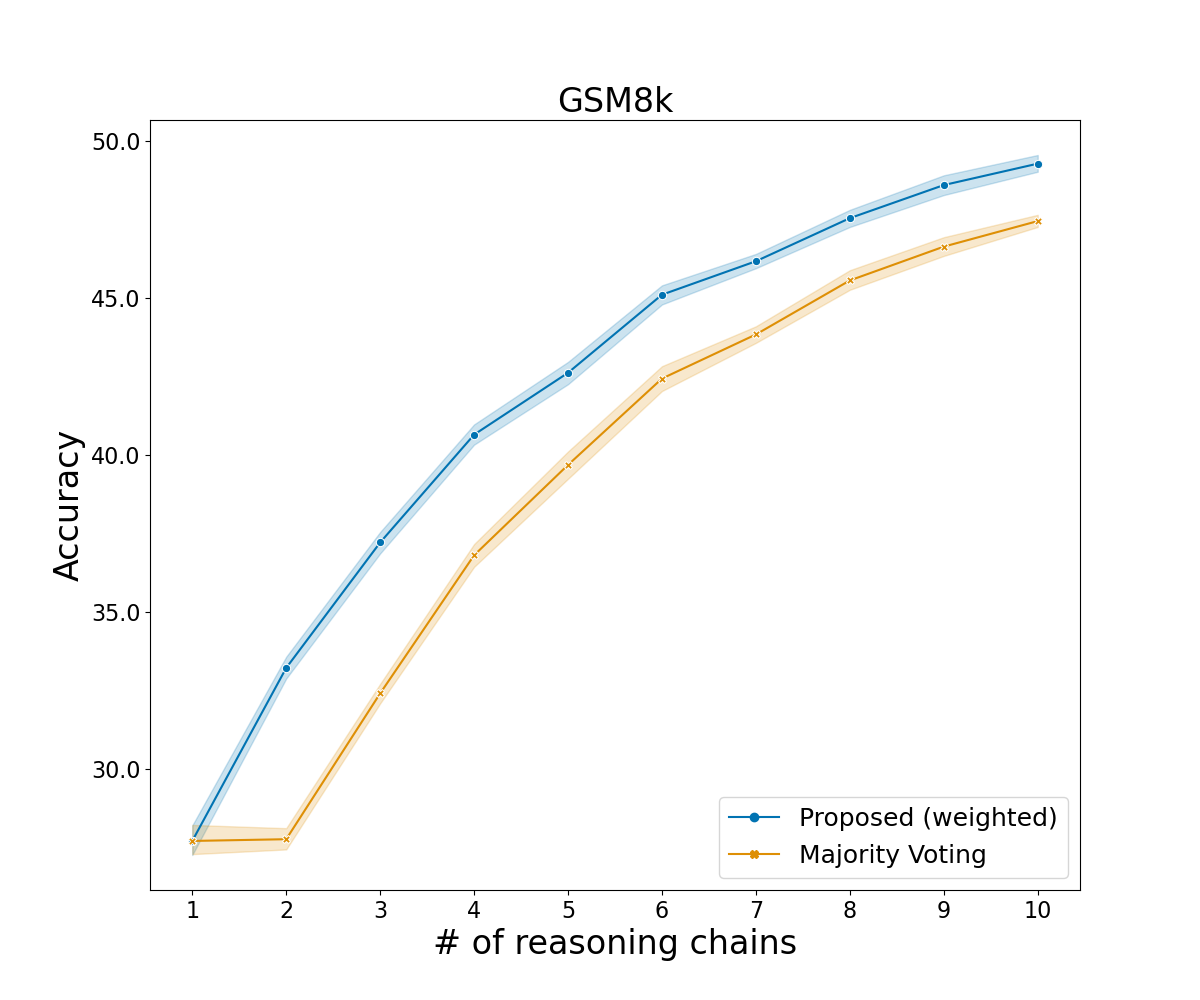}
        \caption{\footnotesize{GSM8k}}
        \label{a:sc_wv_vs_ms:d7}
    \end{subfigure}
    \hfill
    \begin{subfigure}[b]{0.31\textwidth}
        \centering
        \includegraphics[width=0.8\textwidth]{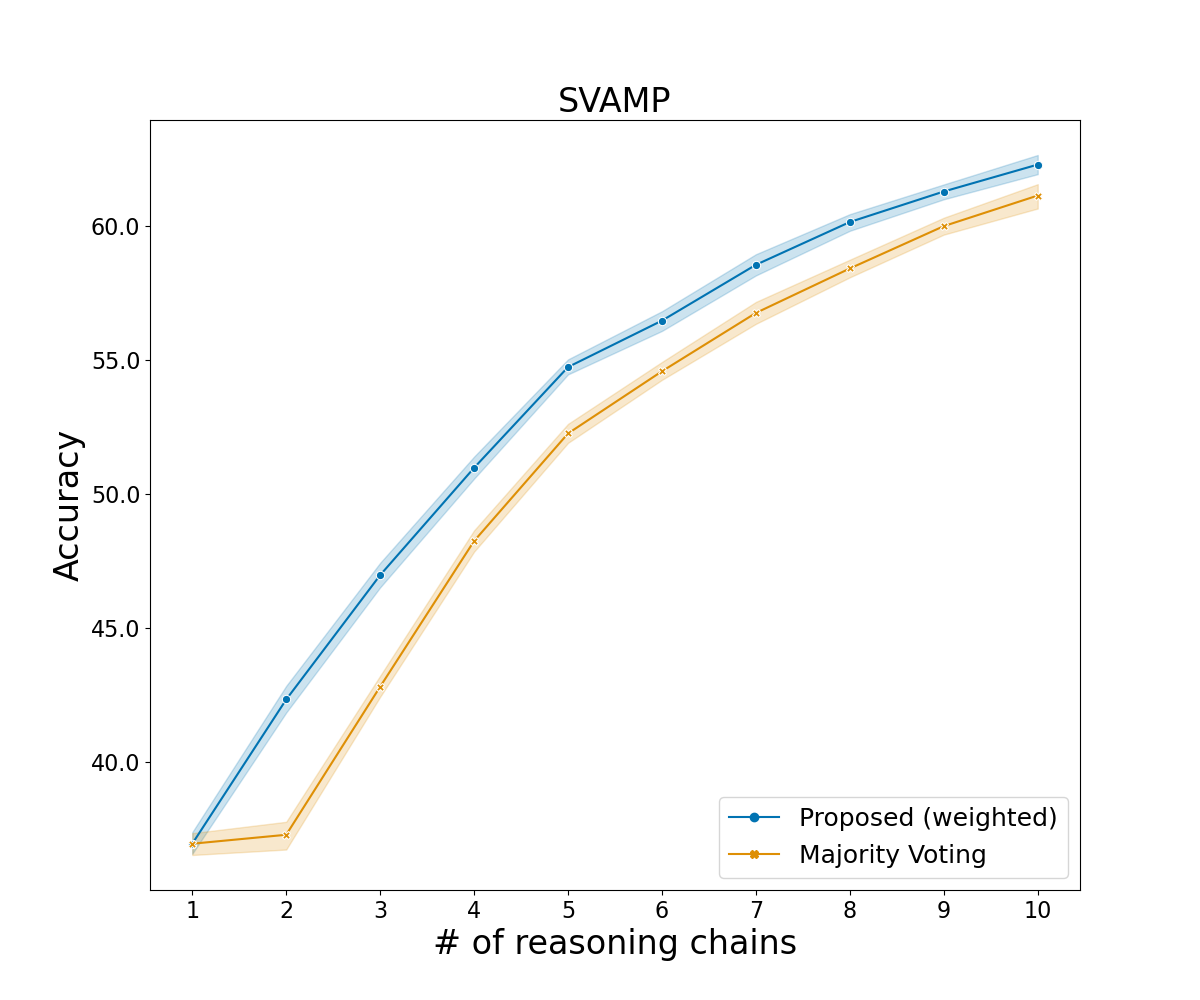}
        \caption{\footnotesize{SVAMP}}
        \label{a:sc_wv_vs_ms:d8}
    \end{subfigure}
    \hfill
    \begin{subfigure}[b]{0.31\textwidth}
        \centering
        \includegraphics[width=0.8\textwidth]{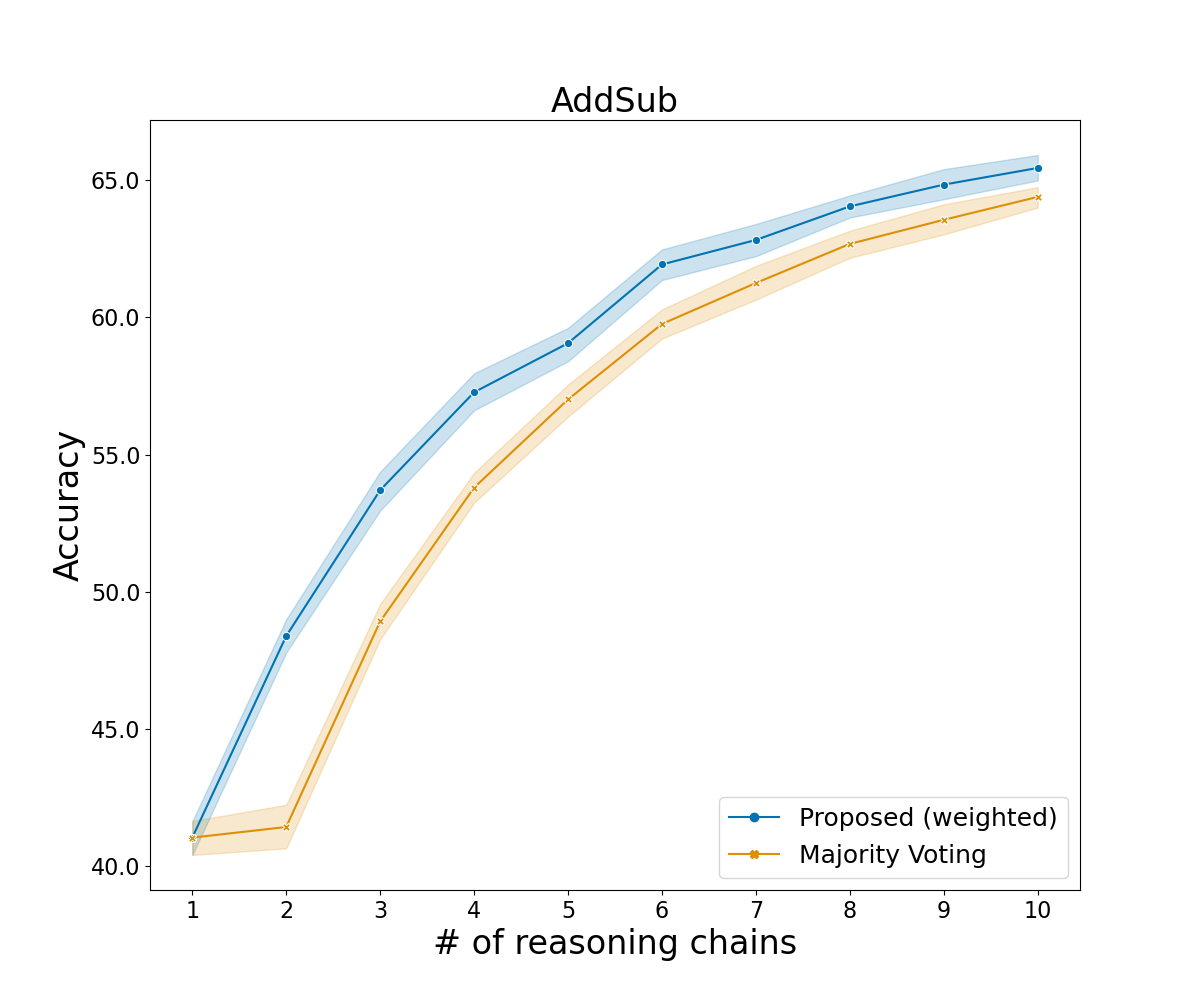}
        \caption{\footnotesize{AddSub}}
        \label{a:sc_wv_vs_ms:d9}
    \end{subfigure}
    \hfill
    \caption{Self-consistency on the \textit{same} reasoning chains, comparing between weighting the final answer using the scores from our proposed verifiers or taking the majority voting ($\uparrow$). Overall, using the scores of our proposed verifiers to perform weighted voting consistently improves the final performance.}
    \label{appendix:fig:sc_wv_vs_mv}    
\end{figure*}

We include in Figure~\ref{appendix:fig:sc_wv_vs_mv} the results over all datasets for the (1) Weighted Voting and (2) Majority Voting over the same reasoning chains.

\section{Ablation}
\label{appendix:ablation}

We include a more comprehensive ablation in Table~\ref{appendix:tab:ablation_sc}, where we ablate over more combinations of the verifiers.

\begin{table*}[t]

    \begin{subtable}[b]{0.95\textwidth}
        \resizebox{0.95\textwidth}{!}{\begin{tabular}{rrrrrrr}
\toprule
\multicolumn{4}{c}{Verifiers} & \multicolumn{3}{c}{Math} \\
\cmidrule(lr){1-4}\cmidrule(lr){5-7}
Perplexity & Relevance & Math Accuracy & Consistency & AddSub & GSM8k & SVAMP \\
\midrule
\xmark & \xmark & \xmark & \xmark & 41.04$\pm$1.79 & 29.23$\pm$1.58 & 38.78$\pm$1.26 \\
\xmark & \xmark & \xmark & \cmark & 48.84$\pm$0.85 & 33.51$\pm$0.69 & 45.41$\pm$0.56 \\
\xmark & \xmark & \cmark & \xmark & 45.76$\pm$1.99 & 35.59$\pm$1.60 & 41.68$\pm$1.65 \\
\xmark & \cmark & \xmark & \xmark & 49.04$\pm$0.44 & 33.38$\pm$0.52 & 46.98$\pm$0.42 \\
\xmark & \cmark & \cmark & \cmark & 53.21$\pm$0.40 & 43.37$\pm$0.25 & 49.83$\pm$0.18 \\
\cmark & \xmark & \xmark & \xmark & 59.09$\pm$0.59 & 41.53$\pm$0.49 & 53.85$\pm$0.45 \\
\cmark & \xmark & \cmark & \xmark & \underline{60.58$\pm$0.47} & 45.66$\pm$0.31 & 55.80$\pm$0.37 \\
\cmark & \xmark & \cmark & \cmark & 59.57$\pm$0.21 & \textbf{48.02$\pm$0.24} & 55.35$\pm$0.24 \\
\cmark & \cmark & \xmark & \xmark & 57.47$\pm$0.33 & 41.87$\pm$0.29 & 54.83$\pm$0.15 \\
\cmark & \cmark & \cmark & \xmark & 59.82$\pm$0.22 & 44.47$\pm$0.22 & \textbf{56.44$\pm$0.24} \\
\cmark & \cmark & \cmark & \cmark & \textbf{62.34$\pm$0.25} & \underline{45.94$\pm$0.30} & \underline{56.36$\pm$0.23} \\
\bottomrule
\end{tabular}
}
        \caption{Ablation Single Chain math results, Accuracy, Higher is better $\uparrow$}
        \label{appendix:tab:ablation_sc_m}
    \end{subtable}
    \hfill\hfill\hfill\hfill\hfill
    \vspace{5mm}
    \begin{subtable}[b]{1\textwidth}
        \resizebox{1.0\textwidth}{!}{\begin{tabular}{rrrrrrrrr}
\toprule
\multicolumn{3}{c}{Verifiers} & \multicolumn{1}{c}{Other} & \multicolumn{3}{c}{Commonsense} & \multicolumn{2}{c}{Symbolic} \\
\cmidrule(lr){1-3}\cmidrule(lr){4-4}\cmidrule(lr){5-7}\cmidrule(lr){8-9}
Perplexity & Relevance & Consistency & BigBench Date & CSQA 2.0 & CSQA & Strategy & Coinflip & Last Letter (2) \\
\midrule
\xmark & \xmark & \xmark & 55.77$\pm$3.06 & 58.75$\pm$1.68 & 47.91$\pm$1.22 & 56.03$\pm$0.99 & 58.67$\pm$2.04 & 15.68$\pm$1.51 \\
\xmark & \xmark & \cmark & 62.70$\pm$0.68 & 56.77$\pm$0.55 & 53.00$\pm$0.59 & 54.31$\pm$0.70 & 49.17$\pm$0.76 & 20.63$\pm$0.78 \\
\xmark & \cmark & \xmark & 62.57$\pm$0.45 & 61.03$\pm$0.21 & 52.62$\pm$0.29 & 57.31$\pm$0.26 & 53.63$\pm$0.51 & 15.79$\pm$0.30 \\
\xmark & \cmark & \cmark & 63.50$\pm$0.26 & 60.47$\pm$0.12 & \underline{54.74$\pm$0.25} & 55.98$\pm$0.24 & 52.68$\pm$0.37 & 16.50$\pm$0.24 \\
\cmark & \xmark & \xmark & 64.06$\pm$0.60 & 60.27$\pm$0.25 & 51.35$\pm$0.36 & \textbf{60.03$\pm$0.30} & \textbf{73.36$\pm$0.42} & \underline{41.73$\pm$0.48} \\
\cmark & \xmark & \cmark & 64.72$\pm$0.31 & 58.74$\pm$0.12 & 52.99$\pm$0.24 & 58.54$\pm$0.13 & 64.89$\pm$0.38 & \textbf{46.03$\pm$0.34} \\
\cmark & \cmark & \xmark & \textbf{71.04$\pm$0.22} & \textbf{62.21$\pm$0.19} & 54.28$\pm$0.14 & \underline{58.98$\pm$0.14} & \underline{69.81$\pm$0.20} & 39.53$\pm$0.21 \\
\cmark & \cmark & \cmark & \underline{69.12$\pm$0.21} & \underline{62.16$\pm$0.22} & \textbf{56.79$\pm$0.12} & 57.21$\pm$0.17 & 64.02$\pm$0.21 & 41.64$\pm$0.44 \\
\bottomrule
\end{tabular}
}
        \caption{Ablation Single Chain non math results, Accuracy, Higher is better $\uparrow$}
        \label{appendix:tab:ablation_sc_nm}
    \end{subtable}
    \caption{Ablation over the types of verifiers used. Overall, all verifiers are meaningfully contributing towards the final solution.}
    \label{appendix:tab:ablation_sc}
\end{table*}

\section{Verifying Incomplete Reasoning Chains}
\label{appendix:incomplete_rc}
\begin{figure*}[!t]

    \begin{subfigure}[b]{0.31\textwidth}
        \includegraphics[width=1.1\columnwidth]{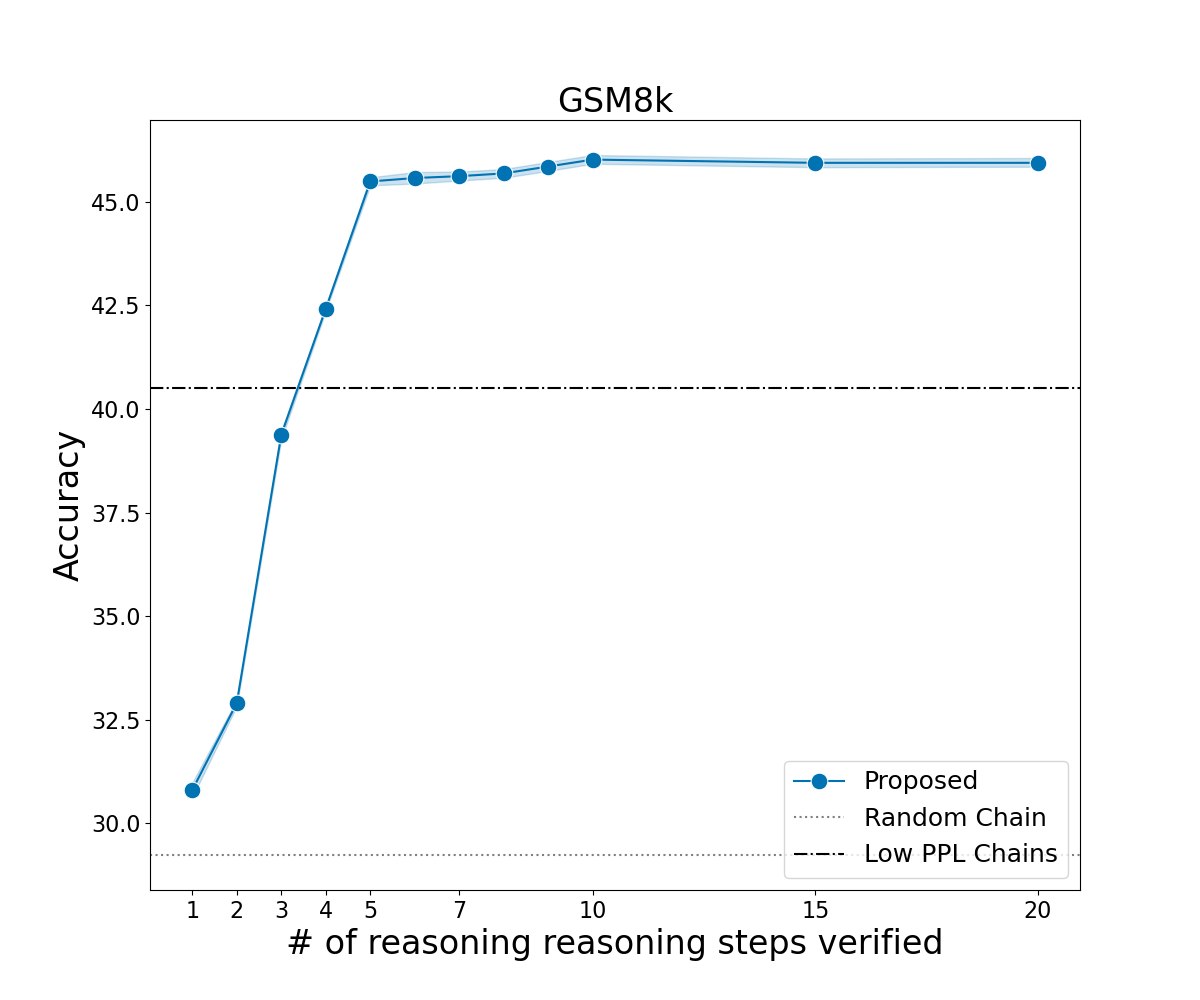}
        \caption{GSM8k}
        \label{l1}
    \end{subfigure}
    \hfill
    \begin{subfigure}[b]{0.31\textwidth}
        \includegraphics[width=1.1\columnwidth]{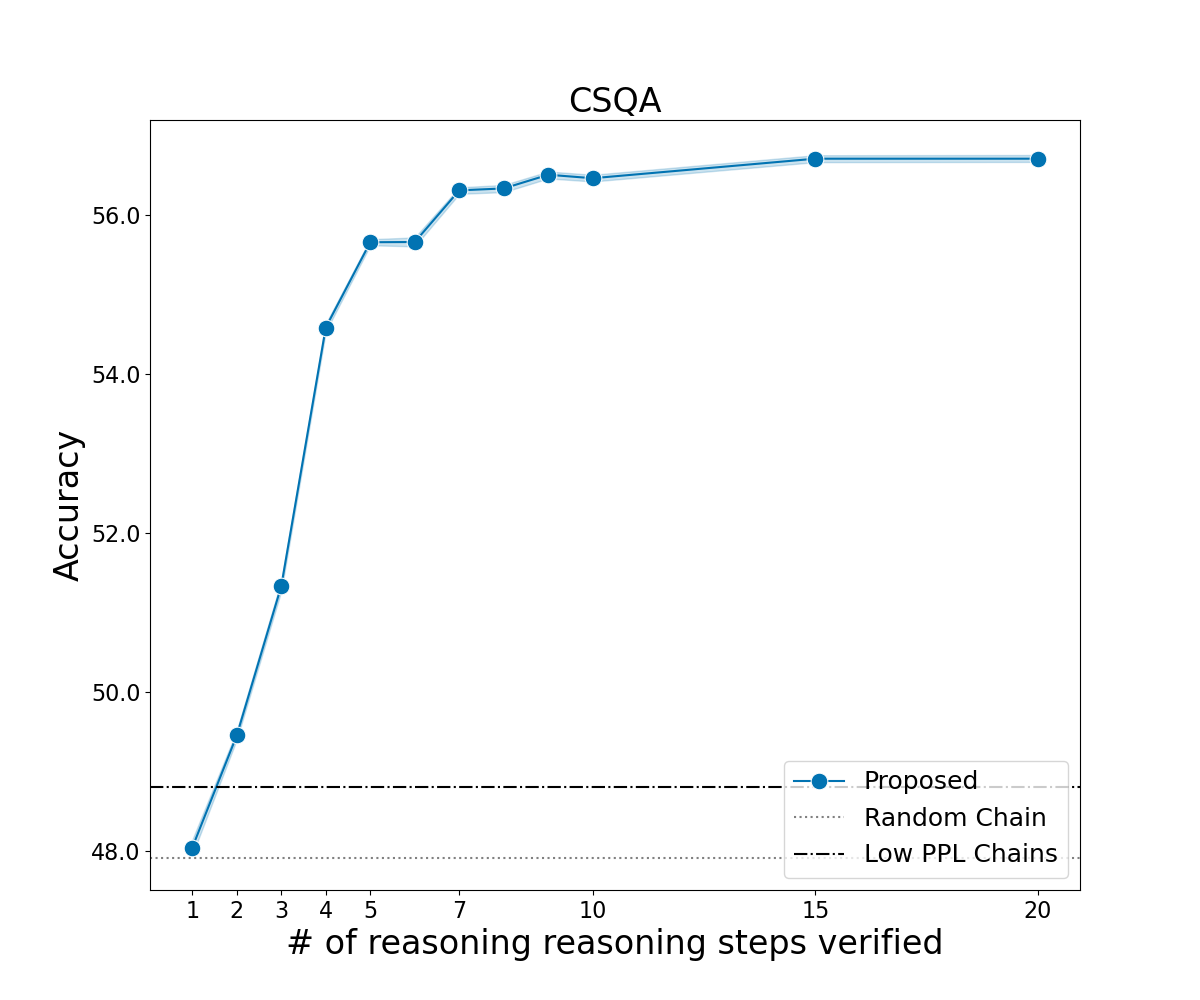}
        \caption{CSQA}
        \label{l1}
    \end{subfigure}
    \hfill
    \begin{subfigure}[b]{0.31\textwidth}
        \includegraphics[width=1.1\columnwidth]{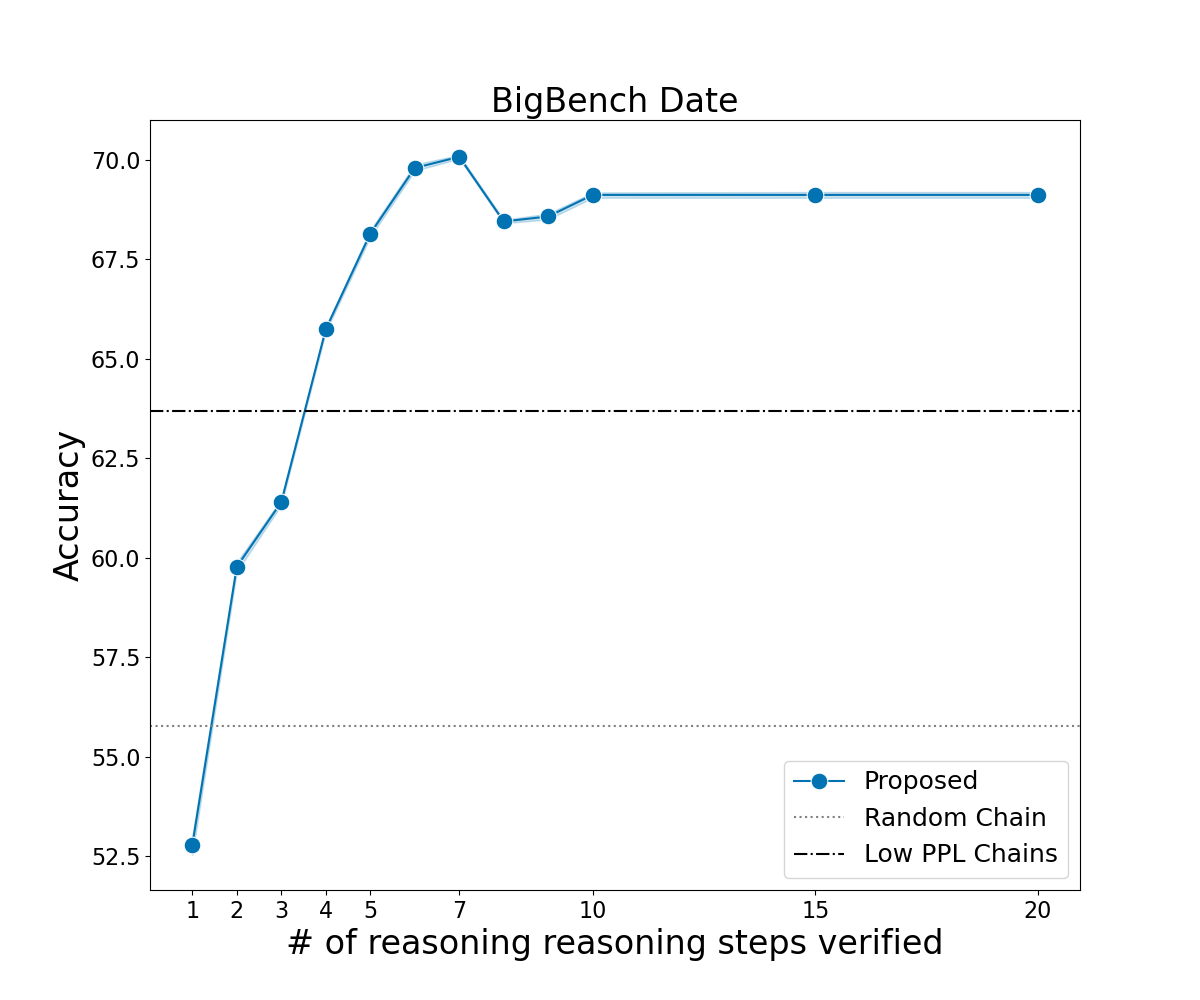}
        \caption{BigBench Date Understanding}
        \label{l1}
    \end{subfigure}
    \caption{Verifying only the first X steps of a chain.}
    \label{appendix:fig:incomplete_rc_abs}
\end{figure*}

To complement Figure~\ref{fig:incomplete_rc_perc}, where we verified a \textit{percentage} of the total number of reasoning steps for a given reasoning chain, we also include Figure~\ref{appendix:fig:incomplete_rc_abs}, where we verify a fixed number of reasoning steps. The motivation behind this experiment is that the number of reasoning steps might not be known beforehand.

To supplement the analysis performed in Section~\ref{ss:incomplete_rc}, we include in Table~\ref{a:tab:incomplete_rc} the performance obtained by our proposed method when verifying a varying number of steps, from 0 (no verification) to All (verify all reasoning steps).

\begin{table*}[h]
    \centering
    \resizebox{1.0\textwidth}{!}{
    \begin{tabular}{rrrrrrrrrrr}
\toprule
& \multicolumn{1}{c}{Other} & \multicolumn{3}{c}{Commonsense} & \multicolumn{2}{c}{Symbolic} & \multicolumn{3}{c}{Math} \\
\cmidrule(lr){2-2}\cmidrule(lr){3-5}\cmidrule(lr){6-7}\cmidrule(lr){8-10}
\cmidrule(lr){2-2}\cmidrule(lr){3-5}\cmidrule(lr){6-7}\cmidrule(lr){8-10}
\# of Steps Verified & BigBench Date & CSQA 2.0 & CSQA & Strategy & Coinflip & Last Letter (2) & GSM8k & SVAMP & AddSub \\
\midrule
0 & 55.77$\pm$3.06             & 58.75$\pm$1.68             & 47.91 $\pm$ 1.22 & 56.03$\pm$0.99             & 58.67$\pm$2.04 & 15.68$\pm$1.51 & 29.23$\pm$1.58 & 38.78$\pm$1.26 & 41.04$\pm$1.79 \\
1 & 52.77$\pm$0.66             & 59.73$\pm$0.26             & 48.04 $\pm$ 0.26 & \textbf{58.65$\pm$0.23}    & 62.03$\pm$0.61 & 40.22$\pm$0.64 & 30.80$\pm$0.53 & 38.12$\pm$0.59 & 46.29$\pm$0.71 \\
2 & 59.77$\pm$0.35             & 61.05$\pm$0.21             & 49.45 $\pm$ 0.16 & 58.22$\pm$0.14             & 67.82$\pm$0.31 & \textbf{51.25$\pm$0.39} & 32.90$\pm$0.24 & 41.31$\pm$0.26 & 50.64$\pm$0.38 \\
3 & 61.40$\pm$0.22             & 61.55$\pm$0.14             & 51.33 $\pm$ 0.17 & 58.04$\pm$0.14             & \textbf{70.60$\pm$0.31} & \underline{46.89$\pm$0.26} & 39.38$\pm$0.29 & 46.27$\pm$0.22 & 51.14$\pm$0.25 \\
4 & 65.75$\pm$0.18             & \textbf{63.40$\pm$0.19}    & 54.58 $\pm$ 0.14 & \underline{58.28$\pm$0.16} & \underline{68.26$\pm$0.36} & 41.49$\pm$0.23 & 42.42$\pm$0.16 & 49.61$\pm$0.20 & 58.02$\pm$0.25 \\
5 & \underline{68.15$\pm$0.25} & \underline{62.35$\pm$0.19} & \underline{55.66 $\pm$ 0.10} & 56.96$\pm$0.17             & 66.72$\pm$0.24 & 33.13$\pm$0.35 & \underline{45.49$\pm$0.26} & \underline{53.38$\pm$0.23} & \underline{60.67$\pm$0.29} \\
All & \textbf{69.12$\pm$0.21}  & 62.16$\pm$0.22             & \textbf{56.79 $\pm$ 0.12} & 57.21$\pm$0.17             & 64.02$\pm$0.21 & 41.64$\pm$0.44 & \textbf{45.94$\pm$0.30} & \textbf{56.36$\pm$0.23} & \textbf{62.34$\pm$0.25} \\
\bottomrule
\end{tabular}
    }
    \caption{Single Chain results, Accuracy, Higher is better $\uparrow$}
    \label{a:tab:incomplete_rc}
\end{table*}

\section{Human Evaluation}
\label{appendix:human_eval}

\subsection{Inter-Annotator Agreement}
\begin{figure*}[t]
    \centering
    \includegraphics[width=0.8\textwidth]{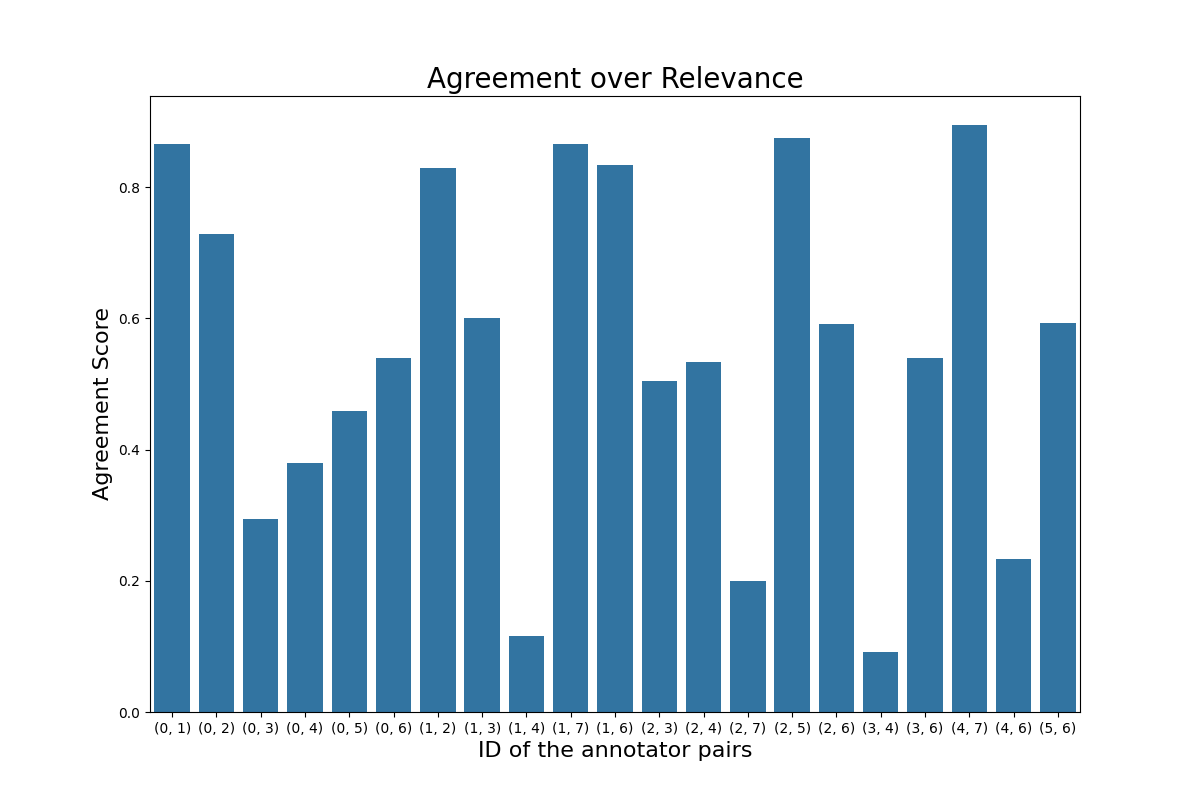}
    \caption{\footnotesize{Annotator Agreement over the Relevance of a given reasoning step}}
    \label{a:ha:agreement_relevance}
\end{figure*}

\begin{figure*}[t]
    \centering
    \includegraphics[width=0.8\textwidth]{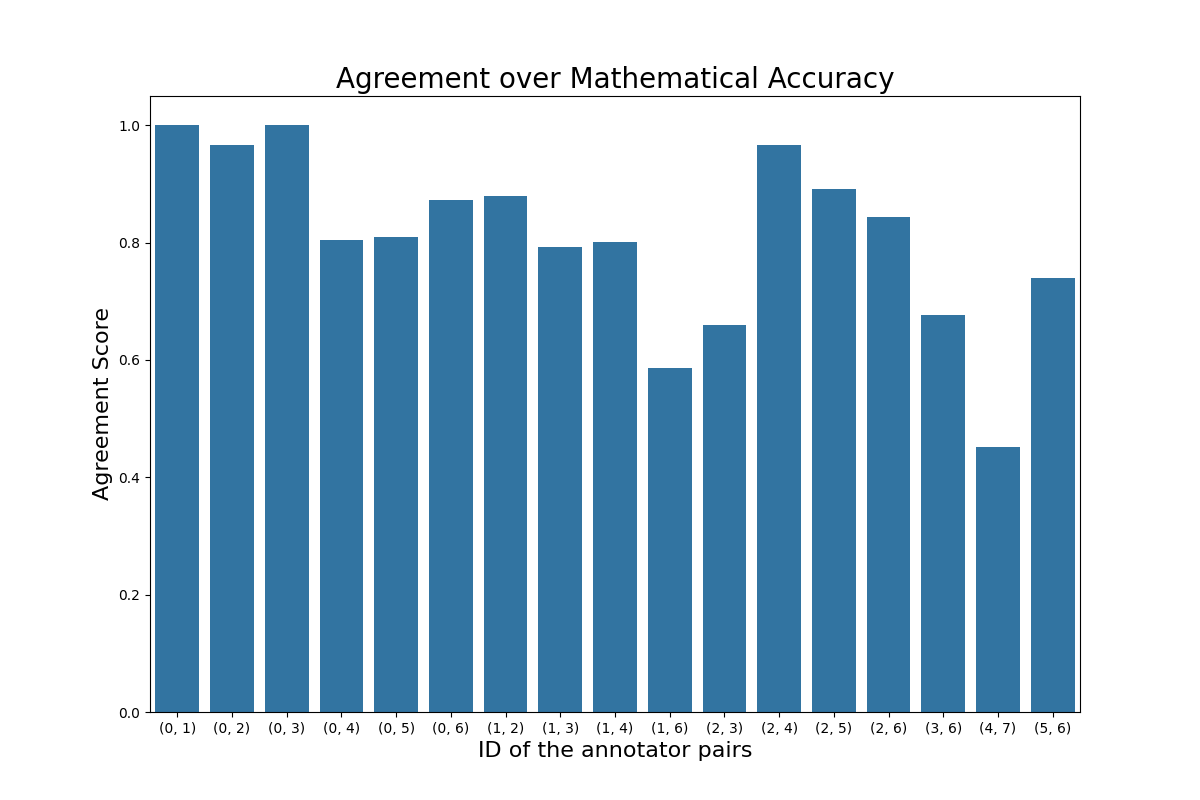}
    \caption{\footnotesize{Annotator Agreement over the Mathematical Accuracy of a given reasoning step}}
    \label{a:ha:agreement_ma}
\end{figure*}

\begin{figure*}[t]
    \centering
    \includegraphics[width=0.8\textwidth]{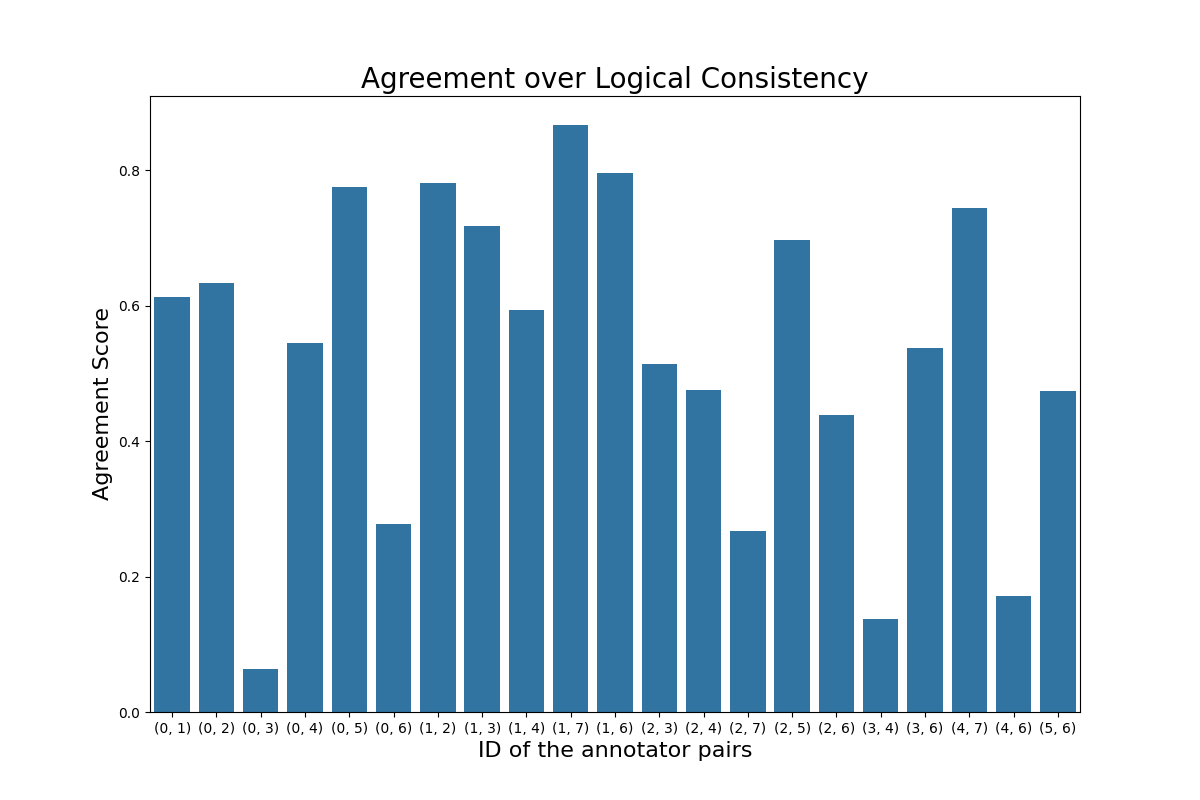}
    \caption{\footnotesize{Annotator Agreement over the Logical Consistency of a given reasoning step}}
    \label{a:ha:agreement_lc}
\end{figure*}

\begin{figure*}[t]
    \centering
    \includegraphics[width=0.8\textwidth]{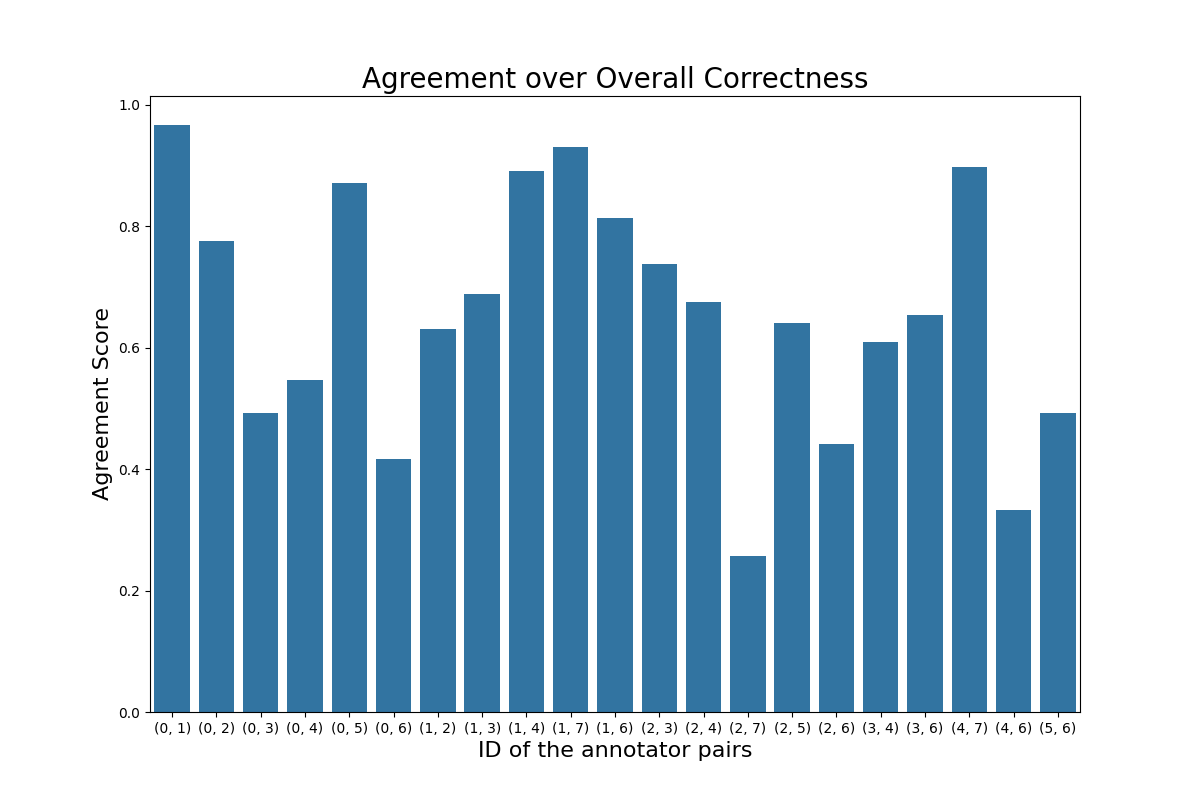}
    \caption{\footnotesize{Annotator Agreement over the Overall Correctness of a given reasoning step}}
    \label{a:ha:agreement_oc}
\end{figure*}

We include the inter-annotator agreement across all 4 measured attributes in Figures~\ref{a:ha:agreement_relevance}~\ref{a:ha:agreement_ma}~\ref{a:ha:agreement_lc}~\ref{a:ha:agreement_oc}. We remark the large variance in the agreement, even for principles that are less subjective (e.g. \textit{Mathematical Accuracy})

\subsection{Correlations}
\begin{figure*}[!t]
    \begin{subfigure}[b]{0.48\textwidth}
        \centering
        \includegraphics[width=0.8\textwidth]{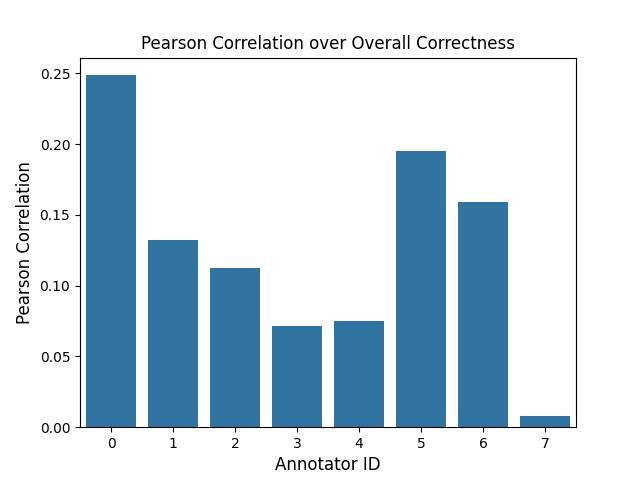}
        \caption{\footnotesize{Correlation score between our overall score (excluding the perplexity verifier) and the annotators assessment.}}
        \label{a:ha:corrleation_oc}
    \end{subfigure}
    \hfill
    \begin{subfigure}[b]{0.48\textwidth}
        \centering
        \includegraphics[width=0.8\textwidth]{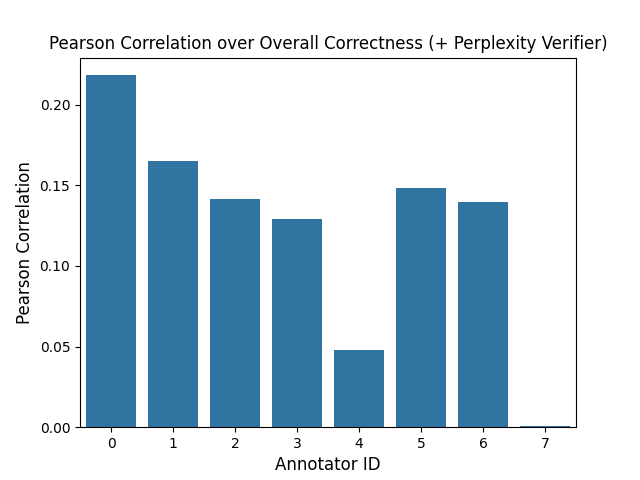}
        \caption{\footnotesize{Correlation score between our overall score (including the perplexity verifier) and the annotators assessment.}}
        \label{a:ha:corrleation_ocppl}
    \end{subfigure}
    \\
    \begin{subfigure}[b]{0.48\textwidth}
        \centering
        \includegraphics[width=0.8\textwidth]{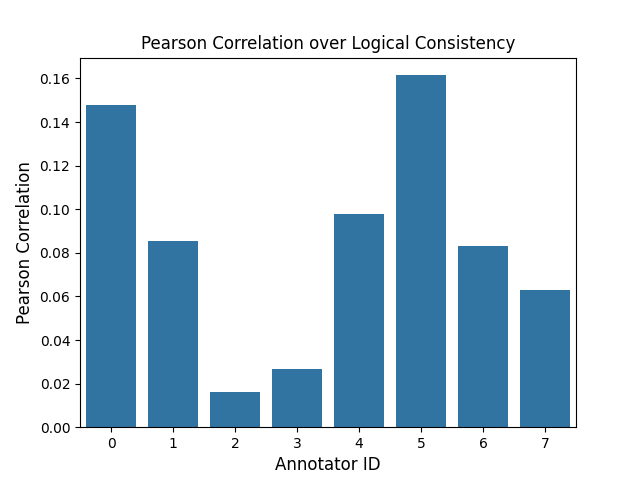}
        \caption{\footnotesize{Correlation score between our Logical Consistency Verifier and the annotators assessment.}}
        \label{a:ha:corrleation_lc}
    \end{subfigure}
    \hfill
    \begin{subfigure}[b]{0.48\textwidth}
        \centering
        \includegraphics[width=0.8\textwidth]{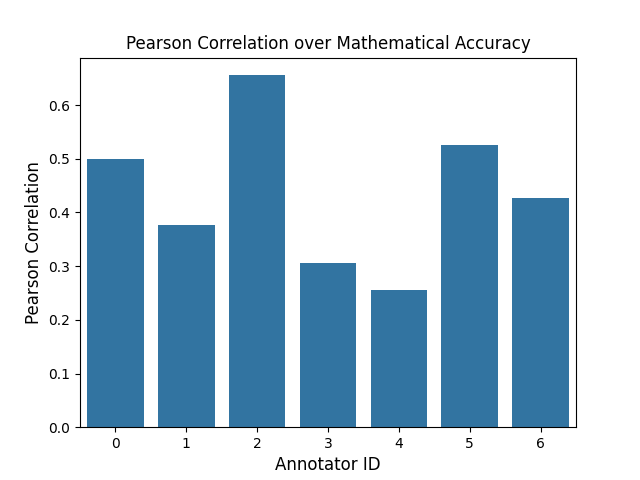}
        \caption{\footnotesize{Correlation score between our Mathematical Accuracy Verifier and the annotators assessment.}}
        \label{a:ha:corrleation_ma}
    \end{subfigure}
    \hfill
    \\
    \begin{subfigure}[b]{0.48\textwidth}
        \centering
        \includegraphics[width=0.8\textwidth]{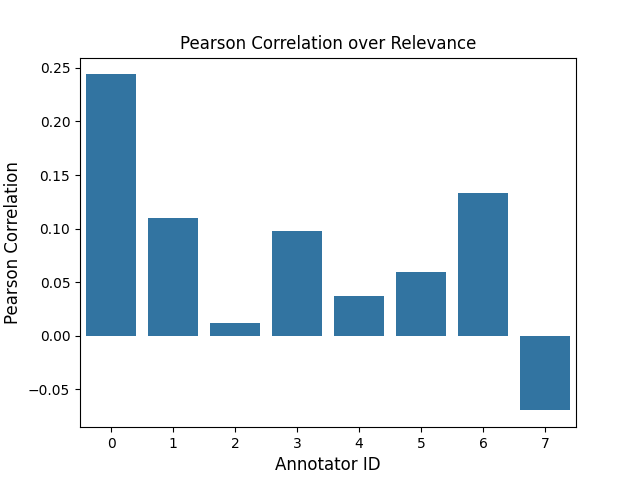}
        \caption{\footnotesize{Correlation score between our Relevance Verifier and the annotators assessment.}}
        \label{a:ha:corrleation_r}
    \end{subfigure}
    \hfill
    \caption{Correlations between the scores of the verifiers and the human annotators' assessment. Additionally, we consider two scores for the \textit{Overall Assessment} attribute. One was computed only with the three verifiers and the second one was computed with the three verifiers, together with the perplexity verifier.}
    \label{appendix:fig:correlations_all_attributes}    
\end{figure*}

We include the Pearson correlations between each of the 4 measured attributes (Relevance, Logical Consistency, Mathematical Accuracy, and Overall Correctness) in Figure~\ref{appendix:fig:correlations_all_attributes}. We remark that there is a large variance in the correlation scores.

\begin{figure}[t]
   \begin{center}
   \includegraphics[width=1.0\columnwidth]{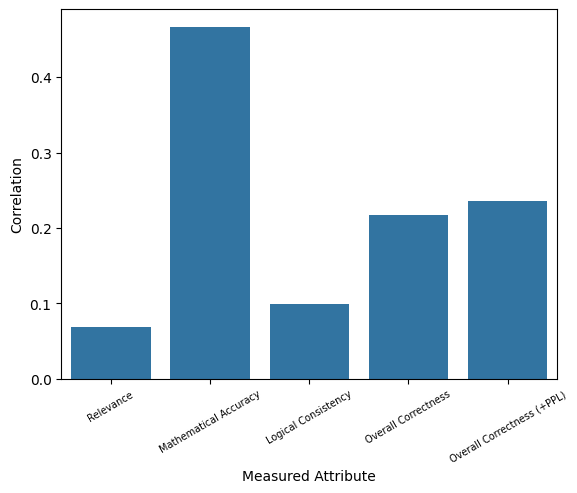}
   \end{center}
   \caption{
    Correlation between the scores of our proposed verifiers and the human assessments. Additionally, we include the correlations between the human assessment of the overall correctness of a given reasoning step and the aggregated score (with and without perplexity).
   }
   \label{a:fig:correlation_verifiers}
\end{figure}

We also include in Figure~\ref{a:fig:correlation_verifiers} the correlations between the verifiers and the human assessments. Additionally, we include the correlations between the human assessment of the overall correctness of a given reasoning step and the aggregated score, with and without perplexity.

\subsection{Human Annotator Instructions}
\label{appendix:ss:human_annotator_instruction}
We provide an overview of the instructions given to the human annotators.

\textbf{(1) Overall Correctness}: If there is any reasoning issue with this step, answer n. If you cannot evaluate the step because you lack expertise or there are some other issues with the step, answer a. Please let us know about the reason in the notes column. If there is nothing wrong with
the reasoning step, answer y.

\textbf{(2) Mathematical Accuracy}: Are all arithmetic calculations in this reasoning step correct? This question is strictly about arithmetic calculations and not about how the
calculation is used to progress toward the solution.

\textbf{(3) Logical Consistency}: Is this reasoning step logically consistent, in itself, and with previous steps? Answer y if the step is logically consistent within itself, and with all previous
steps, including the prompt (problem statement). A step is logically consistent
when it uses available information in a way that is logically correct. In most cases
this means that the conclusions that are reached in this step follow logically from
assumptions made. It can also mean that the step does not contradict
information provided in previous steps (or the same step).

\textbf{(4) Relevance}: Does this reasoning step add information that is relevant for solving the
problem? Information is relevant when it is useful for solving the problem (e.g. it states helping
assumptions, or it reaches a conclusion that answers the problem or get you closer to an
answer). Information can be added by re-stating information from the prompt, by
reaching a conclusion, or by introducing completely new information – any of these can
be relevant or irrelevant.

\section{Expected Performance Given Correlation Levels}
\label{a:s:performance_given_correlation}
We showed in Sections~\ref{ss:single_chain} and \ref{ss:self_consistency} that employing the proposed verifiers leads to performance improvements. Then, we analyzed in Section~\ref{s:he} the correlations between the proposed verifiers and human judgments, observing significantly positive (but low) correlations. 
In this section, we further analyze what improvements can we expect for a given level of correlations. 
To this end, we conducted additional experiments using artificially generated data, allowing us precise control over the correlation values between the verifiers and human judgments. By randomly sampling scores to simulate the verifiers' and human annotators' judgments, we manipulated the data to induce positive correlations and recorded the resulting final scores.\footnote{The correctness of reasoning chains in the artificially generated data is determined based on the sampled scores representing human judgments. A reasoning chain is considered correct if it consists of more than 75\% correct reasoning steps.}

We summarize our results in Table~\ref{a:tab:expected_performance_given_correlation}. We remark that even for modest correlations, the performance increase is over $20\%$ relative, in line with what we observed empirically with real data.

\begin{table}[htbp]
\centering
\begin{tabular}{@{}lr@{}}
\toprule
Method                                           & Score \\ 
\midrule
Model with no verifier                           & 0.19  \\
Model with verifier with $correlation=0.075$     & 0.22  \\
Model with verifier with $correlation=0.1$       & 0.23  \\
Model with verifier with $correlation=0.15$      & 0.26  \\
Model with verifier with $correlation=0.25$      & 0.31  \\
Model with verifier with $correlation=0.5$       & 0.48  \\
Model with verifier with $correlation=0.75$      & 0.72  \\
Model with verifier with $correlation=1.0$       & 0.98  \\ 
\bottomrule
\end{tabular}
\caption{Scores of the proposed method given various level of correlations. Even for modest correlations, the performance increase is over $20\%$ relative.}
\label{a:tab:expected_performance_given_correlation}
\end{table}

\end{document}